\documentclass{article}
\usepackage[paper=a4paper]{geometry}
\usepackage{graphicx} 
\usepackage{amssymb}
\usepackage{pifont}
\usepackage{amsmath}
\usepackage{algorithm}
\usepackage{algpseudocode}
\usepackage{hyperref}
\usepackage[dvipsnames]{xcolor}
\usepackage{colortbl}
\hypersetup{
    colorlinks=false,
    breaklinks=true
}
\usepackage{todonotes}
\usepackage{enumitem}
\usepackage{forest}
\usepackage{longtable}
\usepackage{booktabs}
\usepackage{multirow}
\useforestlibrary{edges}
\usepackage{rotating}
\usepackage{anyfontsize}
\usepackage{floatpag}
\usepackage{float}
\usepackage{tikz}
\usetikzlibrary{positioning, arrows.meta, decorations.pathmorphing, decorations.pathreplacing, shapes.geometric, calc, shadows.blur, backgrounds, fit}
\usepackage{caption}
\usepackage{subcaption}
\usepackage{titling}
\floatpagestyle{plain}

\newcommand{\cmark}{\ding{51}}%
\newcommand{\xmark}{\ding{55}}%

\emergencystretch=\maxdimen
\definecolor{darkgreen}{HTML}{085008}
\definecolor{darkblue}{HTML}{0B3B8D}
\definecolor{darkorange}{HTML}{E46A19}
\definecolor{lightgreen}{HTML}{F0FFF2}
\definecolor{lightblue}{HTML}{F0FCFF}
\definecolor{lightorange}{HTML}{FDE7DA}

\renewcommand{\floatpagefraction}{0.69}

\usepackage[T1]{fontenc}
\usepackage{lmodern}

\usepackage[style=numeric,backend=biber,url=true, doi=true, sorting=none, natbib=true]{biblatex}
\usepackage{bibentry}
\title{FluxDisco: Symbolic Regression for Stoichiometric Dynamical Systems via Monte Carlo Graph Search}
\author{
    \normalsize Cassandra Durr$^{1}$,  Dr Alvaro Köhn-Luque$^{2}$, Prof. Chris Jewell$^{1}$, Dr Lloyd A. C. Chapman$^{1}$ \\
    \small $^{1}$Lancaster University \quad $^{2}$University of Oslo
}

\date{\vspace{-1.2em}}

\begin{document}

\maketitle

\begin{abstract}
    \noindent Dynamical symbolic regression methods identify governing differential equations from noisy data, balancing interpretability and predictive accuracy. However, standard methods often produce expressions that violate known physical laws. To address this, we propose FluxDisco, a physics-informed framework tailored for flux-based, stoichiometric ODE systems. By leveraging a known stoichiometry, we reduce the expression search space and ensure physical adherence. Our framework adapts the Monte Carlo Graph Search algorithm for the unique challenges associated with joint flux discovery of stoichiometric systems. We evaluate our method across a range of physical and biological systems, demonstrating its ability to accurately recover governing dynamics through interpretable equations.
    \\
    
    \\
    
    \noindent\textbf{Code}: \url{https://github.com/CassandraDurr/FluxDisco}
\end{abstract}

\section{Introduction}

Many complex, dynamical systems can be described using a set of ordinary differential equations (ODEs) of the general form
\begin{align*}
    \frac{d\mathbf{x}(t)}{dt} = \dot{\mathbf{x}}(t)=f(\mathbf{x}(t), \theta)
\end{align*}
where $\mathbf{x}(t)$ describes the state of the system at time $t$ and $f$ describes the governing dynamical equations, parametrised by $\theta$. 
This mathematical framework is the basis of dynamical system identification which seeks to build models of dynamical systems using observed data.

Traditionally, researchers have approached the problem of dynamical system identification by assuming the functional form of the governing equations \textit{a priori}, and then using observed data to fit the parameters $\theta$. This `white-box' approach is reasonable if we understand the governing dynamics of the system up to a set of unknown constants. However, assuming such knowledge is often unrealistic for real-world systems so using this approach often entails making restrictive and potentially incorrect assumptions. If the assumed form of $f$ mischaracterises the true dynamics, the misspecified model will inevitably result in inaccurate forecasts and flawed inferences.

In recent years, the scientific community has increasingly turned to `black-box' methods, driven by advances in the field of machine learning, to overcome the limitations of inflexible, `white-box' approaches. 
These data-driven methods learn to forecast the states of dynamical systems, or the derivatives thereof, from historical data without providing an explicit mathematical formulation of the governing dynamics \cite{billings2013nonlinear, chen2018NODE}. 
These frameworks are often very capable emulators of complex dynamics, providing accurate predictions of how dynamical systems may evolve over time. However, their impressive predictive abilities often come at the expense of model interpretability. Building a strong understanding of governing system dynamics is often of equal importance to having accurate predictions. In these cases, purely data-driven methods fall short. Additionally, having interpretable model output often fosters greater confidence in the model.

`Grey-box' modelling provides an opportunity to combine the strengths of traditional `white-box' methods and data-driven `black-box' approaches. Equation discovery can be considered a grey-box approach as it allow us to learn system dynamics directly from observed data without imposing restrictive assumptions, and it outputs interpretable expressions to describe the governing dynamical equations. The class of equation discovery methods we consider in this work is symbolic regression methods. 

Symbolic regression methods can be split into two categories: functional symbolic regression and dynamical symbolic regression. The former has been more widely studied and consists of estimating a function $f(\mathbf{x})$ from paired, input-output data, $(\mathbf{x}, f(\mathbf{x}))$. In contrast, dynamical symbolic regression is used to infer the form of differential equations, $\dot{\mathbf{x}}(t)
= f(\mathbf{x}(t))$, from observed trajectory data $(t, \mathbf{x}(t))$ \cite{d2023odeformer}. 
Dynamical symbolic regression allows us to simultaneously learn the functional form of the ODEs, $f(\mathbf{x}(t), \theta)$, and estimate its parameters, $\theta$.

While pure equation discovery typically imposes no assumptions on the underlying dynamics, we may not be completely oblivious to the governing dynamical laws. This work focuses on \textit{flux-based}, \textit{stoichiometric} systems where the dynamics can be represented as
\begin{align}
    \dot{\mathbf{x}}(t) &= \mathbf{S} \mathbf{v} (\mathbf{x}(t))\label{eq:stoichiometry}.
\end{align}
In this context, $\mathbf{S}$ is a stoichiometry matrix encoding the system's structure and $\mathbf{v}(\mathbf{x}(t))$ is a vector of flux expressions, and we have suppressed dependence of $\mathbf{v}$ on the model parameters $\theta$ for notational convenience, as we do throughout this paper. In physical systems, the underlying stoichiometry is typically known based on prior physical knowledge, whereas the functional form of the fluxes is often unknown. Traditionally, symbolic regression methods fail to account for coupled dynamics in constrained systems, producing solutions that violate known physical laws.
Our work assumes a known stoichiometry and performs \textit{flux discovery}, estimating the system's flux expressions jointly through a novel dynamical symbolic regression framework we call \textit{FluxDisco}.
Rather than performing unconstrained equation discovery, we benefit from imposing a known stoichiometry that reduces the search space of possible expressions and ensures better physical-adherence. 

\subsection{Background}
To discover the governing dynamics of flux-based stoichiometric systems, we treat the generation of candidate flux expressions as a search problem. This section establishes and motivates the framework required to navigate this vast search space efficiently. 

\subsubsection{Markov Decision Process Formulation} \label{sect:MDP}
Sequential decision-making problems involve taking a series of actions where each choice influences long-term outcomes. This class of problems can be formalised using the mathematical framework of finite Markov Decision Processes (MDPs) \cite{sutton2018reinforcement}. An MDP is defined by a set of states $s \in \mathcal{S}$ representing possible environment configurations, and a set of actions $a \in \mathcal{A}(s)$ available from each state. 
As an agent navigates this environment, it receives feedback on its actions through a reward function, $r(s)$. 
The ultimate goal of the agent is to learn a policy for selecting actions that maximises expected cumulative reward.

\begin{figure}
    \centering
    \begin{tikzpicture}[
            state/.style={
                rectangle,
                draw=darkblue,
                fill=lightblue,
                rounded corners,
                text centered,
                minimum height=1.3cm,
                minimum width=1.6cm,
            },
            arrow/.style={
                -Stealth,
                thick,
                draw=darkblue
            }
        ]
        \node[state, align=center] (S) {$x_0(t)$ \\ \textit{(S)}};
        \node[state, align=center, right=2cm of S] (I) {$x_1(t)$ \\ \textit{(I)}};
        \node[state, align=center, right=2cm of I] (R) {$x_2(t)$ \\ \textit{(R)}};
        \draw[arrow] (S) -- (I) node[midway, above] {$v_0(\mathbf{x}(t))$}; 
        \draw[arrow] (I) -- (R) node[midway, above] {$v_1(\mathbf{x}(t))$}; 
    \end{tikzpicture}
    \caption{Flow diagram of the susceptible-infectious-recovered (SIR) epidemic model with state variables $\mathbf{x}(t)$ and fluxes $\mathbf{v}(\mathbf{x}(t))$.} \label{fig:sir_model}
\end{figure}
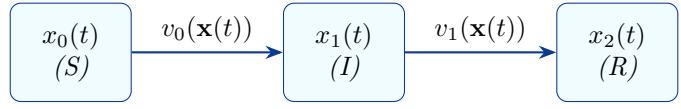

\paragraph{Illustrative Example: SIR Model} 
To bridge the gap between an abstract MDP and the task of flux discovery, we introduce the susceptible-infectious-recovered (SIR) epidemic ODE system, visually represented in Figure~\ref{fig:sir_model}, as a running example. 
The stoichiometry matrix, $\mathbf{S}\in \mathbb{R}^{D\times F}$, and flux vectors, $\mathbf{v}(\mathbf{x}(t)) \in \mathbb{R}^F$, of the SIR system can be represented as,
\begin{align} 
\dot{\mathbf{x}}(t) &= \mathbf{S}\mathbf{v}(\mathbf{x}(t)) \nonumber \\
    \begin{bmatrix}
    \dot{x}_0(t) \\
    \dot{x}_1(t) \\
    \dot{x}_2(t)
    \end{bmatrix} &= \begin{bmatrix}
    -1 & 0\\
    1 & -1\\
    0 & 1
    \end{bmatrix} \begin{bmatrix}
    v_0(\mathbf{x}(t))\\
    v_1(\mathbf{x}(t)) \\
    \end{bmatrix} = \begin{bmatrix}
    -v_0(\mathbf{x}(t))\\
    v_0(\mathbf{x}(t)) - v_1(\mathbf{x}(t)) \\
    v_1(\mathbf{x}(t))
    \end{bmatrix}\label{eq:sir_fluxes}.
\end{align}
The vector $\mathbf{x}(t)=(x_0(t),x_1(t),x_2(t))^\top$ represents the system variables---the susceptible, infectious and recovered population proportions---and $\dot{\mathbf{x}}(t)$ denotes their respective time derivatives. Our objective is to jointly discover flux expressions $\mathbf{v}(\mathbf{x}(t))$. 

\paragraph{MDPs for Flux Discovery}
We define the components of the flux discovery MDP as follows:
\begin{itemize}
    \item \textbf{States} ($s \in \mathcal{S}$): A state represents a set of complete, or partially-complete, flux expressions. We jointly estimate flux expressions, therefore the states include all fluxes undergoing estimation. Flux expressions are composed of terminal tokens (system variables and constants) and non-terminal placeholder tokens ($m$) representing mathematical expressions that require further expansion. Terminal states represent complete expressions consisting of terminal tokens, whereas partial states contain non-terminal tokens.
    For example, a valid state comprising partial expressions for the SIR system (Equation~\ref{eq:sir_fluxes}) is:
    \begin{equation}
    \label{eq:partial_examples}
    \begin{aligned}
        v_0(\mathbf{x}(t)) &= m + m \times x_0(t) \\
        v_1(\mathbf{x}(t)) &= m \times (x_1(t) - c)
    \end{aligned}
    \end{equation}
    where $c$ is a constant token. 
    \item \textbf{Actions} ($a \in \mathcal{A}(s)$): The set of valid actions for a state evolve non-terminal $m$ tokens through the use of a \textit{context-free grammar}. The context-free grammar defines how $m$ can be replaced by operators, constants, or system variables. 
    For example, a grammar for the aforementioned SIR system may be defined as:
    \begin{equation}\label{eq:grammar}
    \begin{aligned}
        m &\to x_0(t) \mid x_1(t) \mid x_2(t) \mid c && \textit{(terminal actions)} \\
        m &\to m+m \mid m-m \mid m \times m \mid \sqrt{m} && \textit{(non-terminal actions)}
    \end{aligned}
    \end{equation}
    \item \textbf{Rewards} ($r(s)$): The reward function evaluates how closely a state's expressions match the observed data.
\end{itemize}

\subsubsection{Monte Carlo Tree Search}
Monte Carlo Tree Search (MCTS) is a popular search algorithm used to learn policies for sequential decision-making \cite{swiechowski2023monte}. The search space is explored through iterative construction of a tree structure, with nodes representing MDP states and the connections between nodes representing valid actions, or grammar rules.
The search algorithm, or agent, intelligently navigates the search space by balancing exploration, trialling unvisited paths, and exploitation, revisiting paths which have demonstrated high potential. 

The algorithm builds up the search tree by cycling through the following phases (illustrated in Figure~\ref{fig:mcts}):
\begin{enumerate}
\item \textbf{Selection}: Starting at the root node, a selection rule is used to traverse the tree until a leaf node is reached.
\item \textbf{Expansion}: If the selected leaf node does not represent a terminal state (complete flux expressions), a new child node is added to the tree.
\item \textbf{Rollout}: Actions (grammar rules) are sampled sequentially from the newly expanded node until a terminal state is reached, yielding a reward. 
\item \textbf{Backpropagation}: The reward obtained is propagated up through the selected nodes, and the estimated values of these nodes are updated to inform future iterations.
\end{enumerate}

\begin{figure}[h!]
    \centering
    \tikzset{
        mcts tree/.style={
            level distance=1.6cm, 
            level 1/.style={sibling distance=1.5cm}, 
            level 2/.style={sibling distance=1.7cm},
            level 3/.style={sibling distance=2cm},
            level 4/.style={sibling distance=2cm},
        },
        root node/.style={
            rectangle,
            rounded corners=3mm,
            draw=darkblue,
            fill=darkblue!70!white,
            text=white,
            very thick,
            solid,
            minimum size=8mm,
            font=\small\bfseries
        },
        highlighted/.style={
            rectangle,
            rounded corners=3mm,
            draw=darkblue,
            fill=lightblue,
            text=darkblue,
            thick,
            solid,
            minimum size=8mm,
            font=\small\bfseries
        },
        normal/.style={
            rectangle,
            rounded corners=3mm,
            draw=darkgray!80,
            fill=gray!10,
            text=darkgray!90,
            thick,
            solid,
            minimum size=8mm,
            font=\small
        },
        highlighted edge/.style={
            very thick,
            color=darkblue,
            ->,
            >=Stealth
        },
        backprop edge/.style={
            very thick,
            color=darkblue,
            ->,
            >=Stealth,
        },
        edge from parent/.style={draw, thick, color=darkgray!90, >=Stealth},
        dag edge/.style={
            thick, color=darkgray!90, >=Stealth, 
        },
    }
    \begin{subfigure}{\textwidth}
        \centering
        \begin{subfigure}[t]{0.24\textwidth}
        \centering
        \resizebox{\linewidth}{!}{
        \begin{tikzpicture}[mcts tree, baseline=(title.north)]
            \node (title) [above=0.5cm, font=\large\bfseries] {Selection};
            \node [style=root node] (root) {$m$}
            child { node [style=normal] {$m \times m$} edge from parent node[left=2pt, font=\footnotesize] {$m \to m \times m$} }
            child { node [style=highlighted] (R1) {$m + m$}
                child { node [style=highlighted] (L2) {$m + m \times m$}
                    child { node [style=normal] {$m + m \times x_0$} 
                        child { node [style=normal] {$m + c \times x_0$} edge from parent node[left=2pt, font=\footnotesize] {$m \to c$} }
                        edge from parent node[left=2pt, font=\footnotesize] {$m \to x_0$}
                    }
                    child { node [style=highlighted] (R3) {$m + m \times c$} 
                    }
                }
                child { node [style=normal] {$m + c$} edge from parent node[right=2pt, font=\footnotesize] {$m \to c$} }
            };
            \draw [style=highlighted edge] (root) -- node[right=2pt, font=\footnotesize, text=darkblue] {$m \to m+m$} (R1);
            \draw [style=highlighted edge] (R1) -- node[left=2pt, font=\footnotesize, text=darkblue] {$m \to m \times m$} (L2);
            \draw [style=highlighted edge] (L2) -- node[right=2pt, font=\footnotesize, text=darkblue] {$m \to c$} (R3);
        \end{tikzpicture}
        }
    \end{subfigure}
    \hfill
    \begin{subfigure}[t]{0.24\textwidth}
        \centering
        \resizebox{\linewidth}{!}{
        \begin{tikzpicture}[mcts tree, baseline=(title.north)]
            \node (title) [above=0.5cm, font=\large\bfseries] {Expansion};
            \node [style=root node] (root) {$m$}
            child { node [style=normal] {$m \times m$} edge from parent node[left=2pt, font=\footnotesize] {$m \to m \times m$} }
            child { node [style=highlighted] (R1) {$m + m$}
                child { node [style=highlighted] (L2) {$m + m \times m$}
                    child { node [style=normal] {$m + m \times x_0$} 
                        child { node [style=normal] {$m + c \times x_0$} edge from parent node[left=2pt, font=\footnotesize] {$m \to c$} }
                        edge from parent node[left=2pt, font=\footnotesize] {$m \to x_0$}
                    }
                    child { node [style=highlighted] (R3) {$m + m \times c$} 
                        child { node [style=highlighted] (D4) {$m + x_0 \times c$} }
                    }
                }
                child { node [style=normal] {$m + c$} edge from parent node[right=2pt, font=\footnotesize] {$m \to c$} }
            };
            \draw [style=highlighted edge] (root) -- node[right=2pt, font=\footnotesize, text=darkblue] {$m \to m+m$} (R1);
            \draw [style=highlighted edge] (R1) -- node[left=2pt, font=\footnotesize, text=darkblue] {$m \to m \times m$} (L2);
            \draw [style=highlighted edge] (L2) -- node[right=2pt, font=\footnotesize, text=darkblue] {$m \to c$} (R3);
            \draw [style=highlighted edge] (R3) -- node[right=2pt, font=\footnotesize, text=darkblue] {$m \to x_0$} (D4);
        \end{tikzpicture}
        }
    \end{subfigure}
    \hfill
    \begin{subfigure}[t]{0.24\textwidth}
        \centering
        \resizebox{\linewidth}{!}{
        \begin{tikzpicture}[mcts tree, baseline=(title.north)]
            \node (title) [above=0.5cm, font=\large\bfseries] {Rollout};
            \node [style=root node] (root) {$m$}
            child { node [style=normal] {$m \times m$} edge from parent node[left=2pt, font=\footnotesize] {$m \to m \times m$} }
            child { node [style=highlighted] (R1) {$m + m$}
                child { node [style=highlighted] (L2) {$m + m \times m$}
                    child { node [style=normal] {$m + m \times x_0$} 
                        child { node [style=normal] {$m + c \times x_0$} edge from parent node[left=2pt, font=\footnotesize] {$m \to c$} }
                        edge from parent node[left=2pt, font=\footnotesize] {$m \to x_0$}
                    }
                    child { node [style=highlighted] (R3) {$m + m \times c$} 
                        child { node [style=highlighted] (D4) {$m + x_0 \times c$} }
                    }
                }
                child { node [style=normal] {$m + c$} edge from parent node[right=2pt, font=\footnotesize] {$m \to c$} }
            };
            \draw [style=highlighted edge] (root) -- node[right=2pt, font=\footnotesize, text=darkblue] {$m \to m+m$} (R1);
            \draw [style=highlighted edge] (R1) -- node[left=2pt, font=\footnotesize, text=darkblue] {$m \to m \times m$} (L2);
            \draw [style=highlighted edge] (L2) -- node[right=2pt, font=\footnotesize, text=darkblue] {$m \to c$} (R3);
            \draw [style=highlighted edge] (R3) -- node[right=2pt, font=\footnotesize, text=darkblue] {$m \to x_0$} (D4);
            \draw [decorate, decoration={snake, amplitude=0.4mm, segment length=2mm, post length=1mm}, ->, >=Stealth, thick, color=darkblue] ($(D4.south)$) -- ++(0,-0.6);
        \end{tikzpicture}
        }
    \end{subfigure}
    \hfill
    \begin{subfigure}[t]{0.24\textwidth}
        \centering
        \resizebox{\linewidth}{!}{
        \begin{tikzpicture}[mcts tree, baseline=(title.north)]
            \node (title) [above=0.5cm, font=\large\bfseries] {Backpropagation};
            \node [style=root node] (root) {$m$}
            child { node [style=normal] {$m \times m$} edge from parent node[left=2pt, font=\footnotesize] {$m \to m \times m$} }
            child { node [style=highlighted] (R1) {$m + m$}
                child { node [style=highlighted] (L2) {$m + m \times m$}
                    child { node [style=normal] {$m + m \times x_0$} 
                        child { node [style=normal] {$m + c \times x_0$} edge from parent node[left=2pt, font=\footnotesize] {$m \to c$} }
                        edge from parent node[left=2pt, font=\footnotesize] {$m \to x_0$}
                    }
                    child { node [style=highlighted] (R3) {$m + m \times c$} 
                        child { node [style=highlighted] (D4) {$m + x_0 \times c$} }
                    }
                }
                child { node [style=normal] {$m + c$} edge from parent node[right=2pt, font=\footnotesize] {$m \to c$} }
            };
            \draw [style=backprop edge] (D4) -- node[right=2pt, font=\footnotesize, text=darkblue] {$m \to x_0$} (R3);
            \draw [style=backprop edge] (R3) -- node[right=2pt, font=\footnotesize, text=darkblue] {$m \to c$} (L2);
            \draw [style=backprop edge] (L2) -- node[left=2pt, font=\footnotesize, text=darkblue] {$m \to m \times m$} (R1);
            \draw [style=backprop edge] (R1) -- node[right=2pt, font=\footnotesize, text=darkblue] {$m \to m+m$} (root);
        \end{tikzpicture}
        }
    \end{subfigure}
        \caption{Monte Carlo Tree Search (MCTS)}
        \label{fig:mcts}
    \end{subfigure}
    \begin{subfigure}{\textwidth}
        \centering
        \vspace{1em}
    \begin{subfigure}[t]{0.24\textwidth}
        \centering
        \resizebox{\linewidth}{!}{
        \begin{tikzpicture}[mcts tree, baseline=(title.north)]
            \node (title) [above=0.5cm, font=\large\bfseries] {Selection};
            \node [style=root node] (root) {$m$}
            child { node [style=normal] {$m \times m$} edge from parent node[left=2pt, font=\footnotesize] {$m \to m \times m$} }
            child { node [style=highlighted] (R1) {$m + m$}
                child { node [style=highlighted] (L2) {$m + m \times m$}
                    child { node [style=normal] {$m + m \times x_0$} 
                        child { node [style=normal] {$m + c \times x_0$} edge from parent node[left=2pt, font=\footnotesize] {$m \to c$} }
                        edge from parent node[left=2pt, font=\footnotesize] {$m \to x_0$}
                    }
                    child { node [style=highlighted] (R3) {$m + m \times c$} 
                    }
                }
                child { node [style=normal] {$m + c$} edge from parent node[right=2pt, font=\footnotesize] {$m \to c$} }
            };
            \draw [style=highlighted edge] (root) -- node[right=2pt, font=\footnotesize, text=darkblue] {$m \to m+m$} (R1);
            \draw [style=highlighted edge] (R1) -- node[left=2pt, font=\footnotesize, text=darkblue] {$m \to m \times m$} (L2);
            \draw [style=highlighted edge] (L2) -- node[right=2pt, font=\footnotesize, text=darkblue] {$m \to c$} (R3);
        \end{tikzpicture}
        }
    \end{subfigure}
    \hfill
    \begin{subfigure}[t]{0.24\textwidth}
        \centering
        \resizebox{\linewidth}{!}{
        \begin{tikzpicture}[mcts tree, baseline=(title.north)]
            \node (title) [above=0.5cm, font=\large\bfseries] {Expansion};
            \node [style=root node] (root) {$m$}
            child { node [style=normal] {$m \times m$} edge from parent node[left=2pt, font=\footnotesize] {$m \to m \times m$} }
            child { node [style=highlighted] (R1) {$m + m$}
                child { node [style=highlighted] (L2) {$m + m \times m$}
                    child { node [style=normal] (L3) {$m + m \times x_0$} edge from parent node[left=2pt, font=\footnotesize] {$m \to x_0$} }
                    child { node [style=highlighted] (R3) {$m + m \times c$} }
                }
                child { node [style=normal] {$m + c$} edge from parent node[right=2pt, font=\footnotesize] {$m \to c$} }
            };
            \node [style=highlighted] (Shared) at ($(L3)!0.5!(R3) - (0, 1.6cm)$) {$m + c \times x_0$};
            \draw [style=dag edge] (L3) -- node[left=2pt, font=\footnotesize, text=darkgray!90] {$m \to c$} (Shared);
            \draw [style=highlighted edge] (root) -- node[right=2pt, font=\footnotesize, text=darkblue] {$m \to m+m$} (R1);
            \draw [style=highlighted edge] (R1) -- node[left=2pt, font=\footnotesize, text=darkblue] {$m \to m \times m$} (L2);
            \draw [style=highlighted edge] (L2) -- node[right=2pt, font=\footnotesize, text=darkblue] {$m \to c$} (R3);
            \draw [style=highlighted edge] (R3) -- node[right=2pt, font=\footnotesize, text=darkblue] {$m \to x_0$} (Shared);
        \end{tikzpicture}
        }
    \end{subfigure}
    \hfill
    \begin{subfigure}[t]{0.24\textwidth}
        \centering
        \resizebox{\linewidth}{!}{
        \begin{tikzpicture}[mcts tree, baseline=(title.north)]
            \node (title) [above=0.5cm, font=\large\bfseries] {Rollout};
            \node [style=root node] (root) {$m$}
            child { node [style=normal] {$m \times m$} edge from parent node[left=2pt, font=\footnotesize] {$m \to m \times m$} }
            child { node [style=highlighted] (R1) {$m + m$}
                child { node [style=highlighted] (L2) {$m + m \times m$}
                    child { node [style=normal] (L3) {$m + m \times x_0$} edge from parent node[left=2pt, font=\footnotesize] {$m \to x_0$} }
                    child { node [style=highlighted] (R3) {$m + m \times c$} }
                }
                child { node [style=normal] {$m + c$} edge from parent node[right=2pt, font=\footnotesize] {$m \to c$} }
            };
            \node [style=highlighted] (Shared) at ($(L3)!0.5!(R3) - (0, 1.6cm)$) {$m + c \times x_0$};
            \draw [style=dag edge] (L3) -- node[left=2pt, font=\footnotesize, text=darkgray!90] {$m \to c$} (Shared);
            \draw [style=highlighted edge] (root) -- node[right=2pt, font=\footnotesize, text=darkblue] {$m \to m+m$} (R1);
            \draw [style=highlighted edge] (R1) -- node[left=2pt, font=\footnotesize, text=darkblue] {$m \to m \times m$} (L2);
            \draw [style=highlighted edge] (L2) -- node[right=2pt, font=\footnotesize, text=darkblue] {$m \to c$} (R3);
            \draw [style=highlighted edge] (R3) -- node[right=2pt, font=\footnotesize, text=darkblue] {$m \to x_0$} (Shared);
            \draw [decorate, decoration={snake, amplitude=0.4mm, segment length=2mm, post length=1mm}, ->, >=Stealth, thick, color=darkblue] ($(Shared.south)$) -- ++(0,-0.6);
        \end{tikzpicture}
        }
    \end{subfigure}
    \hfill
    \begin{subfigure}[t]{0.24\textwidth}
        \centering
        \resizebox{\linewidth}{!}{
        \begin{tikzpicture}[mcts tree, baseline=(title.north)]
            \node (title) [above=0.5cm, font=\large\bfseries] {Backpropagation};
            \node [style=root node] (root) {$m$}
            child { node [style=normal] {$m \times m$} edge from parent node[left=2pt, font=\footnotesize] {$m \to m \times m$} }
            child { node [style=highlighted] (R1) {$m + m$}
                child { node [style=highlighted] (L2) {$m + m \times m$}
                    child { node [style=highlighted] (L3) {$m + m \times x_0$} }
                    child { node [style=highlighted] (R3) {$m + m \times c$} }
                }
                child { node [style=normal] {$m + c$} edge from parent node[right=2pt, font=\footnotesize] {$m \to c$} }
            };
            \node [style=highlighted] (Shared) at ($(L3)!0.5!(R3) - (0, 1.6cm)$) {$m + c \times x_0$};
            \draw [style=backprop edge] (Shared) -- node[left=2pt, font=\footnotesize, text=darkblue] {$m \to c$} (L3);
            \draw [style=backprop edge] (Shared) -- node[right=2pt, font=\footnotesize, text=darkblue] {$m \to x_0$} (R3);
            \draw [style=backprop edge] (L3) -- node[left=2pt, font=\footnotesize, text=darkblue] {$m \to x_0$} (L2);
            \draw [style=backprop edge] (R3) -- node[right=2pt, font=\footnotesize, text=darkblue] {$m \to c$} (L2);
            \draw [style=backprop edge] (L2) -- node[left=2pt, font=\footnotesize, text=darkblue] {$m \to m \times m$} (R1);
            \draw [style=backprop edge] (R1) -- node[right=2pt, font=\footnotesize, text=darkblue] {$m \to m+m$} (root);
        \end{tikzpicture}
        }
    \end{subfigure}
        \caption{Monte Carlo Graph Search (MCGS)}
        \label{fig:mcgs}
    \end{subfigure}
    \caption{The four phases of the MCTS and MCGS algorithm applied to symbolic regression. Unlike MCTS, MCGS allows nodes to be reached via multiple paths, enabling more efficient information propagation across the search structure. For simplicity, a single flux expression per state is illustrated. Highlighted nodes indicate the path traversed during a single iteration of the algorithm. Within-node text shows the flux expression, between-node text shows the applied grammar rule, and arrow directions indicate information flow.}
    \label{fig:mcts-vs-mcgs}
\end{figure}

A disadvantage of tree-based methods is that they fail to identify that different paths can result in identical states. For example, expressions $m+c\times x_0$ and $m+x_0 \times c$ in Figure~\ref{fig:mcts} are mathematically equivalent, but remain distinct states within the tree structure.
Without merging these duplicate states, information can only be backpropagated along the path currently being traversed. 

\subsubsection{Monte Carlo Graph Search}
\citet{leurent2020monte} developed an adaption of MCTS which operates on a graph structure called Monte Carlo Graph Search (MCGS), illustrated in Figure~\ref{fig:mcgs}. This method overcomes the aforementioned limitation of tree-based search algorithms in environments where states can be reached through multiple distinct action sequences. 
By operating directly on a graph, MCGS allows information to flow freely from a node through all of its ancestors to the root node. This global sharing of information sharpens the value estimates of nodes and improves sample efficiency. 

The benefits of a graph-based search are particularly advantageous in environments with a high-degree of state overlap so merging duplicate states dramatically reduces the possible search space. 
The use of a graph is particularly beneficial for symbolic regression due to the nature of mathematical expressions. Because there are multiple ways to construct the same mathematical expression, we expect a high degree of overlap in states. 
Figure~\ref{fig:mcgs} provides an example of how two distinct states can evolve into equivalent expressions that require state merging.

\subsection{Related Work} 
There are three main classes of ODE discovery methods: sparse regression, symbolic regression, and deep learning approaches. \textit{Sparse regression} methods are foundational in this field and include methods such as SINDy (Sparse Identification of Nonlinear Dynamics) and its many variants \cite{brunton2016discovering}. These methods use regularised regression to identify key terms from a user-specified library of candidate functions to form part of the governing dynamical equations. 
In contrast, \textit{symbolic regression} methods are search-based and involve exploring a vast combinatorial space of mathematical operators, variables, and functions to build equations.  
Lastly, \textit{deep learning approaches} include sequence-to-sequence transformers which encode numerical trajectory data and output predicted equations as a sequence of tokens. These methods prioritise generalisability by pre-training on large, synthetic datasets. After pre-training, inference on new ODE systems can be performed almost instantaneously \cite{d2023odeformer, becker2023predicting, vastl2024symformer}.

\paragraph{Symbolic Regression for Stoichiometric Systems}
A small subset of equation discovery methods are tailored towards stoichiometric, dynamical systems. An early example of such an approach is Reactive SINDy \cite{hoffmann2019reactive, jiang2022identification} which learns both the stoichiometry matrix and flux expressions for chemical reaction systems using a user-specified library of possible reactions. This work operates on the assumption of \textit{mass action kinetics} whereby reaction rates are proportional to the product of the concentrations of reactants raised to the power of their stoichiometric coefficients in the balanced chemical equations. This assumption restricts candidate flux expressions to \textit{polynomials}. 
For example, suppose we have a system of $N$ reactions between $M$ chemicals \cite{massactionlecture, wilkinson2018stochastic}, which can be written as:
\begin{align*}
    \sum_{i=1}^M a_{ij} X_i \to \sum_{i=1}^M b_{ij} X_i, \quad j\in \{1,\dots,N\}
\end{align*}
where $(a_{ij})$ and $(b_{ij})$ represent the coefficients of reactants and products $(i=\{1,\dots,M\})$ respectively in reaction $j$, and $X_i$ represents a molecule of the $i^{\text{th}}$ chemical. 
Assuming mass action kinetics, the reaction rates for this system are polynomials,
\begin{align*}
    r_j = k_j \prod_{i=1}^M x_i(t)^{a_{ij}}, \quad j\in \{1,\dots,N\}
\end{align*}
where $(x_i(t))_{i=\{1,\dots,M\}}$ are the numbers of molecules or concentrations of the reactants, with resulting ODEs
\begin{align*}
    \dot{\mathbf{x}}(t) &= \mathbf{S}\mathbf{v}(\mathbf{x}(t))
    \\
    (\mathbf{S})_{ij} &= b_{ij} - a_{ij}
    \\
    (\mathbf{v}(\mathbf{x}(t)))_j &= r_j
    \\
    \dot{x}_i(t) &= \sum_{j=1}^N (b_{ij} - a_{ij})r_j.
\end{align*}
Reactive SINDy requires the specification of candidate reactions, therefore, the algorithm only needs to identify the subset of correct reactions and learn the rates $k_j \,\forall\, j$. 

A significant limitation of Reactive SINDy is the manual pre-specification of candidate reactions. The Stoichiometrically-Informed Symbolic Regression (SISR) method \cite{banos2026stoichiometrically} overcomes this limitation by formulating equation discovery as a search problem, utilising a genetic algorithm to explore the expression space. While our approach is also search-based, we use a grammar-based graph search approach rather than an evolutionary algorithm. Like Reactive SINDy, SISR operates on the assumption of mass action kinetics, restricting the discovered fluxes to polynomials. However, unlike prior methods, SISR discovers the polynomial terms alongside the rate parameters $k_j$ and the stoichiometric matrix $\mathbf{S}$. 

The final equation discovery method for stoichiometric systems that we consider is KinFormer \cite{chen2025kinformer}. This method utilises a transformer architecture and is pre-trained on a large dataset of synthetic chemical reaction systems. KinFormer is similar to the generalised ODEFormer model \cite{d2023odeformer}, although the authors specialise the method for chemical reaction systems. 
While both ODEFormer and KinFormer leverage comprehensive pre-training, ODEFormer considers a wide range of mathematical operators and functions, whereas KinFormer restricts the discovered dynamics to polynomials based on mass action kinetics.

Our philosophy differs fundamentally from the existing equation discovery methods for flux-based, stoichiometric systems. 
While our method assumes a known stoichiometry, it allows for the identification of complex, non-polynomial dynamics.
In contrast, the existing methods are able to uncover stoichiometries, but restrict the discovered dynamics to polynomials based on mass action kinetics.
This assumption is often unrealistic in real-world settings where systems frequently exhibit non-polynomial dynamics.
To our knowledge, among equation discovery methods specifically designed for stoichiometric systems, ours is the first to support non-polynomial flux discovery without relying on the restrictive assumption of mass action kinetics.

\section{Methods}
\subsection{Search Algorithm} \label{sect:search}

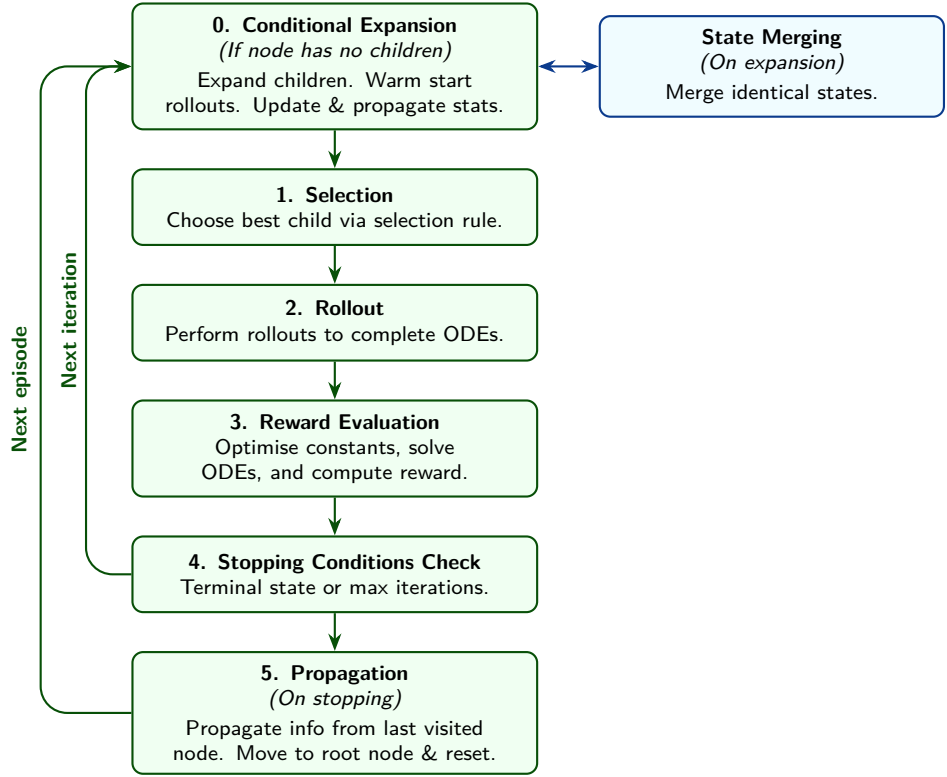
\begin{figure}[htbp]
\centering
\begin{tikzpicture}[
    node distance=0.5cm and 0.8cm,
    font=\footnotesize\sffamily,
    stepnode/.style={
        rectangle,
        rounded corners=4pt,
        draw=darkgreen,
        fill=lightgreen,
        thick,
        align=center,
        inner sep=5pt,
        text width=5cm,
        minimum height=1cm
    },
    tablebox/.style={
        rectangle,
        draw=darkblue,
        thick,
        fill=lightblue,
        align=center,
        inner sep=5pt,
        rounded corners=4pt,
        font=\footnotesize\sffamily, 
        text width=4.2cm,
        minimum height=1cm
    },
    arrow/.style={
        ->, 
        >=Stealth, 
        thick, 
        color=darkgreen,
        rounded corners
    },
    backprop/.style={
        ->,
        >=Stealth,
        thick,
        color=darkgreen,
        rounded corners=10pt
    },
    prop/.style={
        ->,
        >=Stealth,
        thick,
        color=darkblue,
        rounded corners=10pt
    }
]


    \node[stepnode] (expand_warm) {
        \textbf{0. Conditional Expansion}\\
        \textit{(If node has no children)}\\[0.2em]
        Expand children. Warm start rollouts. Update \& propagate stats.
    };
    \node[stepnode, below=of expand_warm] (select) {
        \textbf{1. Selection}\\
        Choose best child via selection rule.
    };
    \node[stepnode, below=of select] (simulate) {
        \textbf{2. Rollout}\\
        Perform rollouts to complete ODEs.
    };
    \node[stepnode, below=of simulate] (evaluate) {
        \textbf{3. Reward Evaluation}\\
        Optimise constants, solve ODEs, and compute reward.
    };
    \node[stepnode, below=of evaluate] (terminal_check) {
        \textbf{4. Stopping Conditions Check}\\
        Terminal state or max iterations.
    };
    \node[stepnode, below=of terminal_check] (propagate) {
        \textbf{5. Propagation}\\
        \textit{(On stopping)}\\[0.2em]
        Propagate info from last visited node. Move to root node \& reset.
    };


    \node[tablebox, right=of expand_warm, anchor=west] (transposition) {
        \textbf{State Merging}\\
        \textit{(On expansion)}\\[0.2em]
        Merge identical states.
    };


    \draw[arrow] (expand_warm) -- (select);
    \draw[arrow] (select) -- (simulate);
    \draw[arrow] (simulate) -- (evaluate);
    \draw[arrow] (evaluate) -- (terminal_check);
    \draw[arrow] (terminal_check) -- (propagate);
    \draw[<->, thick, >=Stealth, darkblue] (expand_warm.east) -- (transposition.west);
    \draw[backprop] (propagate.west) -- ++(-1.2,0) |- (expand_warm.west) 
        node[pos=0.25, rotate=90, anchor=south, font=\footnotesize\sffamily\bfseries] {Next episode};
    \draw[backprop] (terminal_check.west) -- ++(-0.6,0) |- (expand_warm.west)
        node[pos=0.25, rotate=90, anchor=south, font=\footnotesize\sffamily\bfseries] {Next iteration};
\end{tikzpicture}
\caption{Workflow of FluxDisco, our modified MCGS algorithm.} 
\label{fig:mcgs-flow}
\end{figure}

This section describes the technical details of our modified version of the MCGS algorithm, FluxDisco. Our key contribution lies in how we have both adapted and specialised MCGS for the purpose of symbolic regression, specifically joint flux discovery of stoichiometric ODE systems. 

FluxDisco, illustrated in Figure~\ref{fig:mcgs-flow}, provides an iterative procedure to intelligently navigate the expansive search space of governing equations for dynamical systems. In this framework, the search algorithm acts as an `agent' that learns to generate governing dynamical equations by trialling various combinations of mathematical operators, states and constants. Each node in the graph traversed by the agent represents a set of partial or complete flux expressions. The learning is episodic in that the agent starts each cycle of the search at the root node with blank expressions and traverses the nodes of the graph building up the flux expressions through a series of applied grammar rules until it produces complete flux expressions at the terminal states. Within an episode, each iteration follows a structured sequence of selection, simulation, and evaluation to refine the agent's understanding of which functional forms best describe the observed data.

An episode of the algorithm proceeds via the following steps (Figure~\ref{fig:mcgs-flow}), each explained in more detail in the sub-sections below (with additional details provided in the Supplementary Material, Section \ref{sect:supp-search}):
\begin{enumerate}[start=0, label=\textbf{\arabic*}]
    \item \textbf{Conditional Expansion} (Section~\ref{sect:cond-expand}): This pre-selection step determines if the current node, starting with the root node at the beginning of each episode, has existing children. If the node has no children, all valid child nodes are generated and compared against the graph's existing nodes to merge mathematically identical states. To prevent a `cold start' where the search algorithm has no data to differentiate between the children, a set of initial rollouts is performed to establish baseline rewards. 
    \item \textbf{Selection} (Section~\ref{sect:select}): Once child nodes are available, the selection rule is used to traverse the graph from the root towards the most promising terminal states.
    \item \textbf{Rollout} (Section~\ref{sect:rollout}): Rollouts are conducted from selected nodes to complete partial flux expressions.
    \item \textbf{Reward Evaluation} (Section~\ref{sect:eval}): The ODE systems corresponding to rollouts are solved and then evaluated against observed data to produce a set of rewards based on equation parsimony and goodness-of-fit.
    \item \textbf{Stopping Condition Check}: This check determines if the recently evaluated node is terminal, representing complete equations, or if the computational budget for the episode has been exhausted. These conditions trigger information propagation and the end of the episode.
    \item \textbf{Propagation} (Section~\ref{sect:prop}): Reward information learnt during the episode is propagated throughout the graph from the last node visited to all ancestor nodes to inform the next episode's search. 
    After propagation, the next episode begins from the root node.
\end{enumerate}

\subsubsection{Selection} \label{sect:select}

The selection rule ensures the search agent navigates the graph efficiently, identifying promising nodes for expansion. 
The algorithm begins selection from the root node, corresponding to the empty state where fluxes can evolve into any expression supported by the grammar, and traverses the graph towards terminal states. 

Many graph nodes will correspond to partially-complete flux expressions that cannot be directly evaluated. Therefore, we use Monte Carlo rollouts to provide intermediate feedback at non-terminal states, employing the stochastic reward variant of MCGS \cite{leurent2020monte}. To account for reward stochasticity, confidence intervals $(l(s'),u(s'))$ for the mean reward $\mathbb{E}(r|s')$ associated with state $s'$ are constructed (see `\textit{Reward Confidence/Credible Intervals}' paragraph below).
The original MCGS selection rule selects the next state $s'\in \mathcal{S}(s)$ according to
\begin{align}
    s \leftarrow \arg\max_{s' \in \mathcal{S}(s)} u(s') + \gamma \mathcal{U} (s') \label{eq:sampling-rule}
\end{align} 
where $\mathcal{U} (s')$ represents the state's upper value bound, discounted by $\gamma$ (Section~\ref{sect:prop}). 

This selection rule balances local reward, $u(s')$, with future potential reward, $\gamma \mathcal{U} (s')$. The local component of the selection rule prioritises nodes providing the highest average reward. 
While appropriate in most stochastic reward environments, maximising the average reward is unsuitable for equation discovery, where the objective is to identify a single optimal functional form rather than one that performs well on average.
Therefore, we propose the following selection rule:
\begin{align}
    s \leftarrow \arg\max_{s' \in \mathcal{S}(s)}  \left\{ \max \left(u(s'), r^*(s')\right) + \gamma \mathcal{U} (s')  \right\}\label{eq:rmax_selection_rule}
\end{align}
where $r^*(s')$ denotes the maximum reward sampled from state $s'$. The local reward component is now bounded below by the maximum sampled reward, preventing the premature soft-pruning of nodes with high observed performance. 

\paragraph{Reward Confidence/Credible Intervals} 
\citet{leurent2020monte} recommend using the Binary Kullback-Leibler (BKL) divergence to construct the mean reward confidence intervals $(l(s), u(s))$.
However, this formulation is excessively conservative in expansive search spaces like those of equation discovery tasks, resulting in an almost uniform state space search (Supplementary Material Section \ref{sect:supp-reward-cis}).
To encourage faster convergence of the confidence intervals, we derive a new formulation based on the theory underlying Thompson sampling \cite{agrawal2012analysis}. We use the generalisation of Thompson sampling to scalar, bounded rewards to obtain a posterior distribution for the mean empirical reward of a state $s$:
\begin{align*}
    (\Theta|R = \mathbf{r}) &\sim \text{Beta}\left(\alpha=1+\sum_{i=1}^n r_i,\ \beta = 1 + n - \sum_{i=1}^n r_i \right)
\end{align*}
where $\mathbf{r} = (r_1, \dots, r_n)^\top$ denotes the state's sampled rewards.
Under this Bayesian framework, we can derive quantile-based credible intervals in place of confidence intervals for the mean empirical reward
\begin{align}
(l(s),u(s)) &= \left(Q^{-1}_{\alpha/2}(\Theta|R = \mathbf{r}), Q^{-1}_{1-\alpha/2}(\Theta|R = \mathbf{r})\right) \label{eq:beta_credible_intervals}
\end{align}
where $Q^{-1}_p$ denotes the $p$-quantile of the posterior Beta distribution.
While frequentist approaches, such as the BKL framework, generate confidence intervals, credible intervals serve a similar purpose by defining a plausible region for the location of the true mean reward.

\begin{figure}[htbp]
    \centering
    \includegraphics[width=0.9\linewidth]{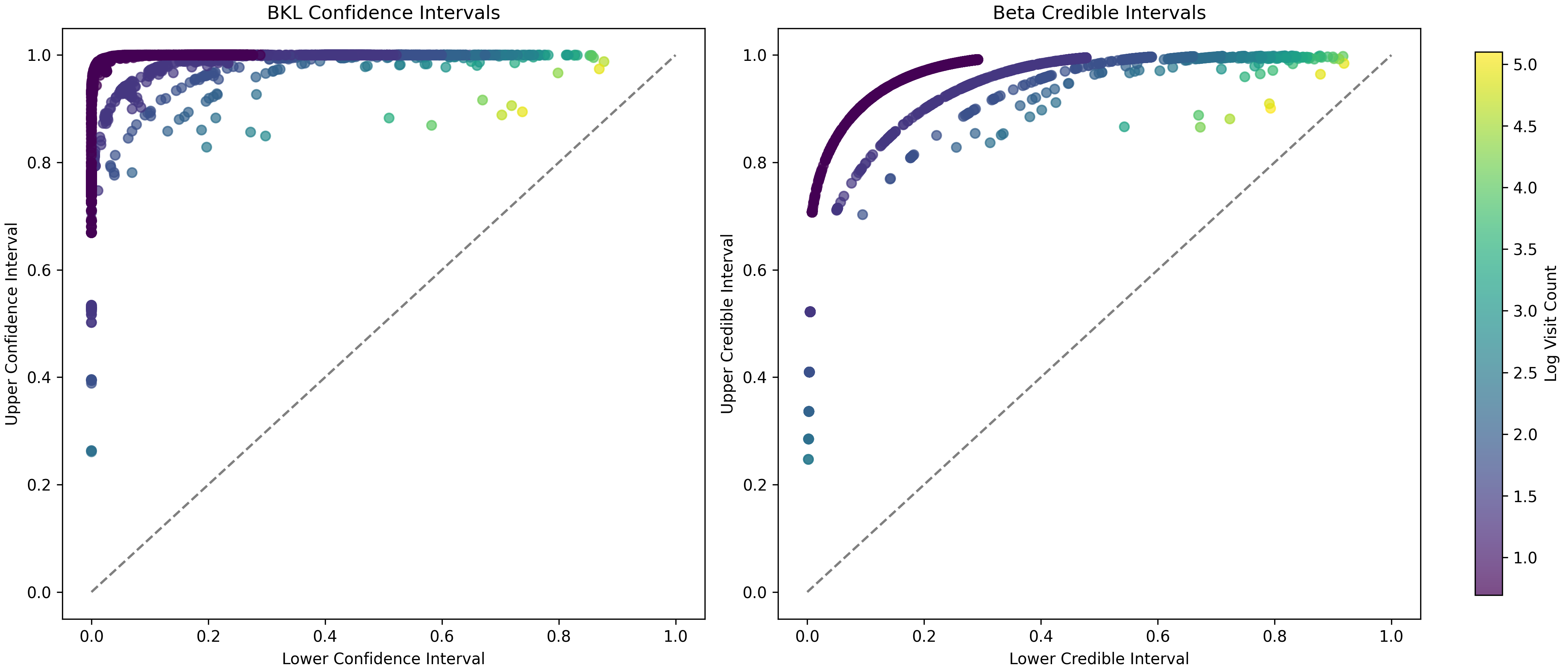}
    \caption{BKL confidence intervals compared with Beta credible intervals (Equation~\ref{eq:beta_credible_intervals}). Both panels illustrate that the more frequently a node is visited, the more its intervals contract towards the reference line, $l(s)=u(s)$.}
    \label{fig:bkl_vs_beta_CI}
\end{figure}

The BKL confidence intervals and Beta credible intervals are compared in Figure~\ref{fig:bkl_vs_beta_CI}.
Most of the BKL upper confidence bounds remain at their initialised value of one, whereas significantly fewer Beta credible interval bounds remain as initialised. 
This behaviour suggests that under the proposed formulation, the sampled rewards are utilised more effectively to differentiate between nodes, allowing for a more targeted search within the expansive state space. 
Having less conservative intervals raises the risk that the true mean falls outside of the intervals, however, this is a worthwhile risk to avoid a uniform exploration of the graph.

\subsubsection{Conditional Expansion} \label{sect:cond-expand}

Before the selection phase can begin, the potential child nodes are created and initialised. Initial rollouts are then performed from each child node to `warm start' their mean reward credible interval bounds $(l(s), u(s))$, and upper and lower value bounds $(\mathcal{L}(s), \mathcal{U}(s))$. 
However, due to the high branching factor associated with symbolic regression, initialising every possible child node introduces significant computational overhead.
As a consequence, we limit the number of child nodes expanded in a single step to the rightmost non-terminal $m$ token for all fluxes. Consequently, multiple iterations may be required to consider all possible children of a state. 

Expansion is further constrained by the hyperparameter $\kappa$, the maximum expression depth.
Following the approach in \cite{sun2022symbolic}, this hyperparameter applies a hard constraint on the complexity of the generated expressions by controlling the number of applied grammar rules per flux. Fluxes with $\kappa$ applied rules have limited expansion---instead of applying all possible grammar rules, only terminal actions are considered.
This constraint concentrates the search on parsimonious expressions, following the inductive bias that physical systems generally comprise few terms \cite{brunton2016discovering, sun2022symbolic}. 
This parsimony constraint is applied in both the expansion phase, and in the rollout phase where fluxes reaching $\kappa$ grammar rules are force-completed using terminal rules. 

Our approach differs from that of \cite{sun2022symbolic}, which awards a zero reward to paths that do not reach a terminal state within $\kappa$ applied rules. In contrast, we opt to complete expressions using terminal rules once $\kappa$ grammar rules have been applied.
The benefit of our approach is that a dense and meaningful reward signal is preserved to better guide the search agent. 
If the grammar allows a wide array of non-terminal actions, \cite{sun2022symbolic}'s method will rarely complete any expressions, leaving most nodes unevaluated and degrading the search. 

\paragraph{State Merging} \label{sect:transposition-table}
To prevent the re-creation of existing graph nodes during expansion, we utilise a transposition table which maps state IDs to the corresponding graph nodes. If the state $s'$ already exists, the parent $s$ is connected to the existing child $s'$, otherwise a new node is created and added to the transposition table. 
The method of serialising states (creating state IDs) should ensure that mathematically equivalent expressions are correctly mapped to the same state ID. To ensure accurate state merging, we apply a \textit{constant folding} procedure. This process simplifies and canonicalises the form of flux expressions, and also reduces the number of constants requiring optimisation. There are two forms of constant folding applied:
\begin{itemize}
    \item \textbf{Additive folding} merges constants which are coefficients to the same terms, e.g., $c_0x_0(t) + c_1x_0(t) \to k_0x_0(t)$.
    \item \textbf{Multiplicative folding} merges constants that are multiplied together, e.g., $c_0c_1x_0(t) \to k_0x_0(t)$.
\end{itemize}
These procedures are applied recursively to equations to catch redundancies within nested sub-expressions, such as root terms or fractions.
Reducing the number of redundant constants aids constant optimisation, improving its stability and increasing the likelihood of finding suitable constants. 
After expressions undergo constant folding, their terms are re-ordered to ensure that terms appear in a consistent, lexicographical order. 

\subsubsection{Rollout} \label{sect:rollout}

Post-selection, independent rollouts are conducted from the chosen node to complete its flux expressions. In completing these expressions, grammar rules are sampled according to predefined probabilities. To incorporate domain-specific knowledge and guide the search towards plausible candidates, we allow user-specified sampling probabilities to be assigned to the grammar rules on a per-flux basis.
These sampling probabilities also provide an additional constraint during the expansion phase; assigning a zero probability to a specific grammar rule prevents the creation of the associated child node.

Grammar rules are repeatedly sampled until either a terminal set of equations is formed, or the fluxes reach $\kappa$ applied grammar rules. If $\kappa$ is reached, the expression undergoes `force-completion', where only terminal rules are sampled (Section~\ref{sect:cond-expand}).

Rollouts provide the opportunity to `probe' the potential of a node. 
After the rollout phase, the rewards associated with completed flux expressions are evaluated to provide samples from the node's underlying reward distribution. These rewards inform a node's credible interval bounds $(l(s), u(s))$ (Section~\ref{sect:select}) and value bounds $(\mathcal{L}(s), \mathcal{U}(s))$. Consequently, nodes with high-reward rollouts will be prioritised during selection, while the value of nodes with poor rollouts are allowed to decay naturally. 

\subsubsection{Reward Evaluation} \label{sect:eval}

During the evaluation phase, the reward for each sampled flux expression is calculated through four main steps:
\begin{enumerate}
    \item \textbf{Constant folding:} To simplify and canonicalise sampled expressions, a constant folding procedure is employed (Section~\ref{sect:transposition-table}). This procedure eliminates redundant constants, simplifying the subsequent constant optimisation process. 
    \item \textbf{ODE system construction:} The simplified expressions are then combined to form a system of ODEs using the pre-specified stoichiometry of the system. 
    \item \textbf{Constant optimisation:} Given the system of ODEs, the unknown constants are optimised iteratively by minimising the error between the observed data and numerical integrations of the ODEs.
    \item \textbf{Reward calculation:} Finally, the reward is calculated using the optimised constants. 
\end{enumerate}
These steps are detailed in Algorithm~\ref{alg:reward_calc} of the Supplementary Material.

The reward calculation is a computational bottleneck of this algorithm due to its use of ODE solving and iterative, multi-dimensional constant optimisation. Many symbolic regression methods avoid ODE solving by using smoothed finite-differencing to estimate state derivatives, which are then used as the search's target. While this approach is more computationally efficient, it has a number of downsides. Finite-differencing greatly amplifies data noise, and while smoothing methods can mitigate this effect, they run the risk of distorting true system dynamics and require the specification of additional hyperparameters. By integrating candidate ODEs, our approach is more robust to noise and preserves the original data as the target of the search. These benefits justify the additional computational expense.

The reward function is designed to balance equation parsimony and goodness-of-fit, and takes the form:
\begin{align}
    R &= \eta^\mathcal{C} \exp \left(- \frac{\text{MSE}_{\text{total}}}{\tau} \right) \label{eq:reward}
\end{align}
where $\eta^\mathcal{C}$ forms the \textit{parsimony penalty} and the exponential term forms the \textit{goodness-of-fit factor}. 

To encourage interpretable solutions, we apply a penalty for non-parsimonious equations, $\eta^\mathcal{C}$ \cite{sun2022symbolic}. The hyperparameter $\eta \ (\le 1)$ is the parsimony coefficient and determines the degree to which complex equations should be penalised. The variable $\mathcal{C}$ represents the system's complexity, where the complexity of a single flux is defined as the number of operations and substitutions required to build its mathematical representation. 
The system's complexity is taken to be the maximum complexity of the individual fluxes:
\begin{align*}
    &\mathcal{C}_i = \text{complexity}(\text{flux }i), \ i \in \{1, \dots, F\}
    \\
    &\mathcal{C} = \max(\mathcal{C}_1, \dots, \mathcal{C}_F).
\end{align*}
For reference, Figure~\ref{fig:complexity} illustrates the complexity calculation for a simple flux.

The flux complexities are calculated after constant folding to reflect the simplified, final forms of the expressions. 
The larger the value of $\mathcal{C}$, or the lower the value of $\eta$, the greater the applied penalty.

\begin{figure}
    \centering
    \begin{tikzpicture}[
        nonterminal/.style={
            circle,
            draw=darkblue,
            fill=lightblue,
            thick,
            solid,
            minimum size=6mm,
            font=\footnotesize\bfseries,
            inner sep=0pt
        },
        terminal/.style={
            circle,
            draw=darkorange,
            fill=lightorange,
            thick,
            solid,
            minimum size=6mm,
            font=\footnotesize\bfseries,
            inner sep=0pt
        },
        operation/.style={
            circle,
            draw=darkgray!80,
            fill=gray!10,
            thick,
            solid,
            minimum size=6mm,
            font=\footnotesize\bfseries,
            inner sep=0pt
        },
        state_container/.style={
            rectangle,
            draw=gray!30,
            fill=gray!5, 
            rounded corners,
            inner sep=5pt,
            minimum width=3.5cm
        },
        arrow/.style={
            -Stealth,
            thick,
            gray!60
        },
        math_label/.style={
            font=\scriptsize\itshape,
            color=darkgray,
        },
    ]
        \node[state_container] (i22) at (0, 0) {
            \begin{tikzpicture}[level distance=0.8cm, sibling distance=1.5cm]
                \node[operation] {$+$}
                    child {node[operation] {$\times$}
                        child {node[nonterminal] {$m$}
                            child {node[terminal] {$x_0$}}
                        }
                        child {node[nonterminal] {$m$}
                            child {node[terminal] {$x_1$}}
                        }
                    }
                    child {node[nonterminal] {$m$}
                        child {node[terminal] {$c$}}
                    };
            \end{tikzpicture}
        };
        \node (h_i22) [above=0.3cm of i22] {Flux: $(x_0(t)\times x_1(t)) + c$};
        \node[draw=none, fill=none, align=left, anchor=north west] (ops) at ([xshift=0.8cm]i22.north east) {
            \textbf{Operations:} \\[0.2em]
            \tikz[baseline=-0.4ex]\node[nonterminal]{$m$}; \tikz[baseline=-0.4ex]\node[operation, draw=none, fill=none] {$+$}; \tikz[baseline=-0.4ex]\node[nonterminal]{$m$}; \\[0.2em]
            \tikz[baseline=-0.4ex]\node[nonterminal]{$m$}; \tikz[baseline=-0.4ex]\node[operation, draw=none, fill=none] {$\times$}; \tikz[baseline=-0.4ex]\node[nonterminal]{$m$};
        };
        \node[draw=none, fill=none, align=left, anchor=north west] (subs) at ([xshift=0.5cm]ops.north east) {
            \textbf{Substitutions:} \\[0.2em]
            \tikz[baseline=-0.4ex]\node[nonterminal]{$m$}; \tikz[baseline=-0.4ex]\node[operation, draw=none, fill=none] {$\leftarrow$}; \tikz[baseline=-0.4ex]\node[terminal]{$c$}; \\[0.2em]
            \tikz[baseline=-0.4ex]\node[nonterminal]{$m$}; \tikz[baseline=-0.4ex]\node[operation, draw=none, fill=none] {$\leftarrow$}; \tikz[baseline=-0.4ex]\node[terminal]{$x_1$}; \\[0.2em]
            \tikz[baseline=-0.4ex]\node[nonterminal]{$m$}; \tikz[baseline=-0.4ex]\node[operation, draw=none, fill=none] {$\leftarrow$}; \tikz[baseline=-0.4ex]\node[terminal]{$x_0$};
        };
        \path (ops.north) -- (subs.north) coordinate[midway] (mid_cols);
        \node (h_c) [above=0.3cm of mid_cols] {Flux complexity = 5};
    \end{tikzpicture}
    \caption{Illustration of the complexity calculation of a flux, where the flux expression is visualised as an expression tree. \textbf{\textcolor{darkblue}{Blue}} nodes represent non-terminal tokens, \textbf{\textcolor{darkorange}{orange}} nodes represent applied terminal actions, and \textbf{\textcolor{gray!90}{grey}} nodes represent applied non-terminal actions.}
    \label{fig:complexity}
\end{figure}
 
The exponential term of the reward function, 
$$\exp \left(- \frac{\text{MSE}_{\text{total}}}{\tau} \right),$$
quantifies the goodness-of-fit between observations and the candidate solution. 
The total mean squared error ($\text{MSE}_{\text{total}}$) can be decomposed as:
\begin{align*}
    \text{MSE}_{\text{total}} &= \sum_{d =0}^{D-1} \left(\frac{1}{N\times T} \sum_{n=1}^N \sum_{i=1}^T \left(\hat{x}_{d}^{(n)}(t_i) - x_{d}^{(n)}(t_i)  \right)^2\right)
\end{align*}
where $\hat{x}_{d}^{(n)}(t_i)$ denotes the model's estimate for state $x_d$ at time $t_i$ for realisation $n$, and $x_{d}^{(n)}(t_i)$ reflects the corresponding observed data. 

The temperature hyperparameter $\tau$ scales the error to the range of the data, preventing a flat and uninformative reward landscape. Without this scaling factor, the goodness-of-fit component of the reward may be unable to distinguish between solutions, allowing the parsimony penalty to dominate the reward function. 

\paragraph{Constant Optimisation} \label{sect:const-opt}

The reward function evaluates candidate ODEs by comparing their integrated trajectories to observed data. Since symbolic constant tokens in the expressions prevent direct numerical integration, the reward is calculated iteratively by proposing constants, integrating the ODEs, and then computing the reward.

We optimise constants in the flux expressions using non-linear least squares estimation. We use the trust region reflective algorithm for this on account of its robustness which is essential when evaluating a broad range of candidate ODEs \cite{trustregion, li1994convergence}. The constant optimisation process is repeated using multiple initialisations run in parallel to avoid getting stuck in local minima of the error landscape.

Candidate expressions are evaluated by numerically integrating systems using the fourth-order Runge–Kutta method (RK4). To promote numerical stability, predicted trajectories are bounded to a range wider than the true data range. 
This prevents trajectories from exploding towards infinite values, causing integration errors. Furthermore, allowing the trajectories to extend beyond the range of the data improves the stability of the constant optimisation procedure, as informative gradients are maintained even for unsuitable constants. 

\subsubsection{Propagation} \label{sect:prop}

Graph propagation occurs at the end of an episode once a terminal state is reached, or a maximum number of iterations have been performed. During this phase, the upper and lower value bounds $(\mathcal{L}(s), \mathcal{U}(s))$ of nodes are updated according to the modified Bellman equations,
\begin{align} 
\mathcal{U}(s) &\leftarrow \max_{s' \in \mathcal{S}(s)} \left\{ \max \left(u(s'), r^*(s')\right) + \gamma \mathcal{U}(s') \right\}  \label{eq:updated_bellman_u}\\
\mathcal{L}(s) &\leftarrow \max_{s' \in \mathcal{S}(s)} \left\{ l(s') + \gamma \mathcal{L}(s') \right\}, \label{eq:updated_bellman_l}
\end{align}
which are constructed to align with the selection rule (Equation~\ref{eq:rmax_selection_rule}).
The bounds are only ever tightened as per the monotonicity condition laid out in \cite{leurent2020monte}.

Propagation begins at the last node visited during the episode and propagates through all ancestor nodes, including parent nodes not visited during the episode. By updating all paths stemming from a visited node, sample efficiency is improved.
The bounds of terminal states, representing complete expressions, collapse to the true reward to reflect the elimination of uncertainty:
\begin{align*}
    \mathcal{L}(s)=r(s)=\mathcal{U}(s).
\end{align*}

To manage the high computational cost of updating bounds over the full graph, we employ an efficient, queue-based implementation of the bound updates proposed by \citet{leurent2020monte} (\textit{Algorithm 4}). While the authors recommend propagation after every iteration in an episode, we perform propagation once at the end of each episode. The drawback of this approach is that node selection is based on slightly outdated information within an episode. However, this adaption significantly reduces the total number of updates. Empirically, this approach provides a significant reduction in computational cost with no noticeable performance degradation.

\subsection{Simulation Study} \label{sect:study-design}

We evaluate the performance of our framework, FluxDisco, in identifying governing equations from simulated data across a variety of flux-based dynamical systems.

To rigorously test our method, we benchmark it against simulated data from a range of systems exhibiting complex characteristics such as fast and slow dynamics, bifurcations, and oscillations.
The benchmarked systems selected for this evaluation are described in Table~\ref{tab:ode_systems}. 
Observations from real-world systems are typically noisy, therefore we simulate data for each ODE system under noiseless, low noise, and high noise conditions. Further details of the data simulation process can be found in Section~\ref{sec:supp-data} of the Supplementary Material.

\begin{sidewaystable}
\centering
\caption{Details of the coupled ODE systems considered in this study, including their stoichiometries $\mathbf{S}$, differential equations $\dot{\mathbf{x}}$, selected grammar rules, and grammar rule exclusions via sampling probabilities.}
\label{tab:ode_systems}
\begin{tabular}{@{}lcclll@{}}
\toprule
\textbf{System} &
  \textbf{$\mathbf{S}$} &
  \textbf{$\dot{\mathbf{x}}$} &
  \textbf{True flux expressions} &
  \textbf{Grammar} &
  \textbf{Exclusions} \\ \midrule
SIR &
  $\begin{bmatrix} -1 & 0\\1 & -1\\0 & 1\end{bmatrix}$ &
  $\begin{bmatrix}-v_0\\v_0 - v_1 \\v_1\end{bmatrix}$ &
  $\begin{aligned}
    v_0 &= \begin{cases}
    c_0x_0x_1, \ \textit{(standard)} \\
    c_0x_0x_1^2, \ \textit{(squared)} \\
    c_0x_0\sqrt{x_1}, \ \textit{(square root)}
    \end{cases} \\
    v_1 &= c_1x_1
    \end{aligned}$ 
    &
  $\begin{aligned} &\text{Terminal actions:} \\ &m \to x_0 \mid x_1 \mid x_2 \mid c  \\
  \ &\text{Non-terminal actions:}
  \\&m \to m+m \mid m-m \mid m \times m \mid \sqrt{m} \end{aligned}$ &
  $\begin{aligned}p(m \to x_2|v_0)&=0 \\p(m \to x_0|v_1)&=0\end{aligned}$ \\ \addlinespace \midrule
  Lotka-Volterra &
  $\begin{bmatrix}
            1 & -1 & 0 \\
            0 & 1 & -1
        \end{bmatrix}$ &
  $\begin{bmatrix}
        v_0 - v_1 \\
        v_1 - v_2
        \end{bmatrix}$ &
  $\begin{aligned}
    v_0 &= c_0x_0 \\
    v_1 &= c_1x_0x_1 \\
    v_2 &= c_2x_1
    \end{aligned}$ 
    &
  $\begin{aligned} &\text{Terminal actions:} \\ &m \to x_0 \mid x_1 \mid c  \\
  \ &\text{Non-terminal actions:}
  \\&m \to m+m \mid m-m \mid m \times m \end{aligned}$ &
  $\begin{aligned}p(m \to x_1|v_0)&=0 \\p(m \to x_0|v_2)&=0\end{aligned}$ \\ \addlinespace \midrule
  Brusselator &
  $\begin{bmatrix}
            1 & -1 & 1\\
            -1 & 1 & 0
        \end{bmatrix}$ &
  $\begin{bmatrix}
        v_0 - v_1 + v_2 \\
        -v_0 + v_1
        \end{bmatrix}$ &
  $\begin{aligned}
    v_0 &= x_0^2x_1 \\
    v_1 &= c_0x_0 \\
    v_2 &= c_1 - x_0
    \end{aligned}$ 
    &
  $\begin{aligned} &\text{Terminal actions:} \\ &m \to x_0 \mid x_1 \mid c \\
  \ &\text{Non-terminal actions:}
  \\&m \to m+m \mid m-m \mid m \times m \end{aligned}$ &
  $\begin{aligned}p(m \to x_1|v_1)&=0 \\p(m \to x_1|v_2)&=0\end{aligned}$ \\ \addlinespace \midrule
  Fairen-Velarde &
  $\begin{bmatrix}
            -1 & 1 & 0 \\
            -1 & 0 & 1
        \end{bmatrix}$ &
  $\begin{bmatrix}
        -v_0 + v_1 \\
        -v_0 + v_2
        \end{bmatrix}$ &
  $\begin{aligned}
    v_0 &= \frac{x_0x_1}{1+c_0x_0^2} \\
    v_1 &= c_1 - x_0\\
    v_2 &= c_2
    \end{aligned}$ 
    &
  $\begin{aligned} &\text{Terminal actions:} \\ &m \to x_0 \mid x_1 \mid c \\
  \ &\text{Non-terminal actions:}
  \\&m \to m+m \mid m-m \mid m \times m \mid \frac{m}{m} \end{aligned}$ &
  \\ \bottomrule
\end{tabular}
\end{sidewaystable}

\paragraph{SIR}
The first system we consider is the SIR epidemic model introduced in Section~\ref{sect:MDP}. 
We consider three different disease transmission behaviours to test the ability of our method to recover different functional forms of the fluxes. To incorporate prior knowledge of the system, we exclude specific grammar rules from being sampled via zeroed sampling probabilities. The grammar rule $m \to x_2(t)$ is excluded from the flux $v_0$, representing the infection process, as the proportion of recovered individuals $x_2(t)$ does not impact the rate of new infections. Similarly, $m \to x_0(t)$ is excluded from flux $v_1$ representing the recovery process, as the proportion of susceptible individuals $x_0(t)$ has no impact on disease recovery. Outside of these exclusions, the remaining sampling probabilities are balanced uniformly.

\paragraph{Lotka-Volterra}
The Lotka-Volterra system is a well-studied ecological model that describes how populations of interacting predator and prey groups fluctuate over time, producing oscillatory behaviours.
In this formulation, the states $x_0(t)$ and $x_1(t)$ represent the population density of the prey and predators, respectively. 
The grammar rule $m \to x_1(t)$ was excluded from $v_0$, as prey birth rates are independent of predator population density, and $m \to x_0(t)$ was excluded from $v_2$, as the natural death rate of predators is independent of prey availability. 

\paragraph{Brusselator}
The Brusselator is a theoretical chemical reaction system which models the concentration of interacting chemicals $x_0(t)$ and $x_1(t)$ whose reaction is driven by constant inputs $c_0$ and $c_1$:
\begin{itemize}
    \item Flux $v_0$ represents an \textit{autocatalytic reaction} that consumes $x_0(t)$ and $x_1(t)$ to produce more $x_0(t)$.
    \item Flux $v_1$ represents a reaction between chemical $x_0(t)$ and input $c_0$ which outputs $x_1(t)$.
    \item Flux $v_2$ represents two reactions: the creation of $x_0(t)$ from input $c_1$, and the removal, or decay, of $x_0(t)$. While this composite term could be split into separate fluxes, combining them reduces the dimensionality of the search.
\end{itemize}

The Brusselator exhibits a \textit{Hopf bifurcation} at the threshold $c_0 = 1 + c_1^2$.
When $c_0 < 1 + c_1^2$, the system stabilises towards a fixed point, and when $c_0 > 1 + c_1^2$, the system produces oscillations characterised by an attracting limit cycle \cite{engel2026singular}. Additionally, this system exhibits \textit{fast and slow dynamics} \cite{engel2026singular}. In the unstable, oscillatory regime, there are phases where the chemical concentrations spike rapidly and others where the concentrations change more gradually.

To capture both dynamical behaviours, we considered constants from both the stable and the unstable regimes. The grammar rule $m \to x_1(t)$ was excluded from fluxes $v_1$ and $v_2$ as the chemical $x_1(t)$ is not a reactant in the corresponding reactions.

\paragraph{Fairen-Velarde}

The Fairen-Velarde model describes an oscillatory bacterial respiration system \cite{fairen1979time}. The dynamical governing equations capture the amount of oxygen, $x_0(t)$, and nutrients, $x_1(t)$, in the system over time.
Unlike previous systems considered, the Fairen-Velarde system includes a rational polynomial flux, $v_0$, which can prove challenging for symbolic regression methods. Furthermore, the functional form of the rational polynomial is significantly more complex than previously considered fluxes, with a minimum complexity of $\mathcal{C}=11$. To compensate for the increased system complexity, we assume that the functional form of the oxygen and nutrient supplies, $v_1$ and $v_2$, are known \textit{a priori}. Therefore, we aim to fully discover the functional form of the consumption flux, $v_0$, and only estimate the constants $[c_1, c_2]$ from fluxes $v_1$ and $v_2$. This assumption is reasonable as the oxygen and nutrient supplies are typically controlled and monitored in an experimental setup. 

\subsubsection{Performance Evaluation} \label{sect:performance_metrics}

To measure the performance of symbolic regression methods, generated expressions can be evaluated \textit{symbolically} or \textit{numerically}. Symbolic evaluation compares the ground truth mathematical expressions with the top-scoring, candidate expressions. 
This evaluation can indicate whether the primary objective of symbolic regression, discovering dynamics from data, is met. 
In comparison, numerical evaluation involves comparing the noiseless trajectory data with the integrated candidate ODE systems, $\hat{\mathbf{x}}(t)$. We utilise both numerical and symbolic evaluation to holistically evaluate model performance.

We also consider computational runtime and efficiency in our evaluation where efficiency measures the balance between computational expense and the predictive error of generated expressions. This measure standardises the comparison between methods by quantifying whether models make great sacrifices in speed or accuracy in favour of the other.

\paragraph{Numerical Evaluation} We evaluate our method's performance quantitatively by computing the normalised mean squared error (NMSE),
\begin{align*}
    \text{NMSE}(\hat{\mathbf{x}}(t), \mathbf{x}_{\text{true}}(t)) &= \frac{100}{T \times \sigma^2_{\text{true}}}\sum_{i=1}^T\left(\mathbf{x}_{\text{true}}(t_i)-\hat{\mathbf{x}}(t_i) \right)^2
    = 100 \frac{\sum_{i=1}^T\left(\mathbf{x}_{\text{true}}(t_i)-\hat{\mathbf{x}}(t_i) \right)^2}{\sum_{i=1}^T\left(\mathbf{x}_{\text{true}}(t_i)-\bar{\mathbf{x}}_{\text{true}} \right)^2}
\end{align*}
between the true, noiseless trajectories $\mathbf{x}_{\text{true}}$ and our estimates $\hat{\mathbf{x}}(t)$ (Figure~\ref{fig:nmse}). 
The NMSE measures the model's error relative to a baseline error obtained by predicting the true data's mean $\bar{\mathbf{x}}_{\text{true}}$ across all time points. 
If the method na\"ively predicts the mean of the true data across all time steps, then a NMSE of 100 is attained. We anticipate that well-performing symbolic regression methods should fit the data at least as well as the constant mean $\bar{\mathbf{x}}_{\text{true}}$, however, the NMSE can grow very large if the predictions are highly inaccurate.

\paragraph{Symbolic Evaluation} Because our work focuses on coupled, stoichiometric systems, we assess ability to identify the coupled expressions of each system (Table~\ref{tab:coupled-terms}). This evaluation is more lenient than full-system identification, as it does not account for uncoupled expressions and does not penalise the inclusion of incorrect terms.

\begin{table}[]
\centering
\caption{Coupled expressions within each ODE system considered}
\label{tab:coupled-terms}
\begin{tabular}{@{}lcc@{}}
\toprule
\textbf{ODE System} & $\dot{\mathbf{x}}$ & \textbf{Number of coupled expressions} \\ \midrule
SIR (Standard) & $\begin{bmatrix}
    \textcolor{darkblue}{-c_0 x_0 x_1} \\ \textcolor{darkblue}{c_0x_0x_1} \textcolor{darkorange}{- c_1x_1} \\ \textcolor{darkorange}{c_1x_1}
\end{bmatrix}$ & 2 \\ \addlinespace
SIR (Squared) & $\begin{bmatrix}
    \textcolor{darkblue}{-c_0 x_0 x_1^2} \\ \textcolor{darkblue}{c_0x_0x_1^2} \textcolor{darkorange}{- c_1x_1} \\ \textcolor{darkorange}{c_1x_1}
\end{bmatrix}$ & 2 \\ \addlinespace
SIR (Square root) & $\begin{bmatrix}
    \textcolor{darkblue}{-c_0 x_0 \sqrt{x_1}} \\ \textcolor{darkblue}{c_0x_0\sqrt{x_1}} \textcolor{darkorange}{- c_1x_1} \\ \textcolor{darkorange}{c_1x_1}
\end{bmatrix}$ & 2 \\ \addlinespace
Lotka-Volterra & $\begin{bmatrix}
    c_0x_0 \textcolor{darkblue}{- c_1x_0x_1} \\ \textcolor{darkblue}{ c_1x_0x_1} - c_2x_2
\end{bmatrix}$ & 1 \\ \addlinespace
Brusselator & $\begin{bmatrix}
    \textcolor{darkblue}{x_0^2x_1} \textcolor{darkorange}{- c_0x_0} + c_1 - x_0 \\ \textcolor{darkblue}{-x_0^2x_1} \textcolor{darkorange}{+ c_0x_0}
\end{bmatrix}$ & 2 \\ \addlinespace
Fairen-Velarde & $\begin{bmatrix}
    \textcolor{darkblue}{- \frac{x_0x_1}{1+c_0x_0^2}} +c_1 - x_0 \\ \textcolor{darkblue}{- \frac{x_0x_1}{1+c_0x_0^2}} + c_2
\end{bmatrix}$ & 1\\ \bottomrule
\end{tabular}
\end{table}

To determine if our method is able to identify a coupled expression, we evaluate its top predicted equations using the reward function (Equation~\ref{eq:reward}) as the criterion for selecting top-performing equations. If the coupled term is present in the relevant state derivatives and the coefficients of this term are sufficiently close, the term is correctly identified. 

\paragraph{Efficiency Evaluation}
To facilitate a fair comparison of the tradeoff between speed and accuracy, we define a composite efficiency metric,
\begin{align*}
    \text{Efficiency} = \frac{1}{\text{Runtime}^{0.5} \times  \text{NMSE}^{0.5}}
\end{align*}
which assigns equal weight to computational expense and predictive error. This metric penalises both methods that achieve marginal accuracy gains at great computational costs, and those that run quickly but fail to produce accurate solutions.

\subsubsection{Method Comparison}

To contextualise our results, we benchmark our approach against a diverse range of popular equation discovery methods using the evaluation metrics outlined in Section~\ref{sect:performance_metrics}. Key characteristics of the benchmarked methods are detailed in Table~\ref{tab:methods}.
Of the methods considered, ours is the only method that accounts for known stoichiometries and estimates system fluxes as opposed to the system ODEs. To the best of our knowledge, there are no existing dynamical symbolic regression approaches which allow a stoichiometry to be imposed. While Reactive SINDy \cite{hoffmann2019reactive, jiang2022identification}, SISR \cite{banos2026stoichiometrically}, and KinFormer \cite{chen2025kinformer} allow stoichiometries and flux expressions to be estimated, they are confined to the restrictive assumption of mass action kinetics which does not hold for all of the ODE systems considered in this study. Furthermore, these methods estimate stoichiometries as opposed to imposing them.
In the absence of directly comparable physics-informed methods for flux-based symbolic regression, we benchmark our approach against a wide range of top-performing generalised approaches. To ensure a fair comparison against these baselines, we evaluate two variants of our framework: our standard stoichiometry-aware method (denoted `Ours') and an ablation without imposed stoichiometries (`Ours (No Stoich.)'). In the Supplementary Materials, Section~\ref{sect:supp-ablation}, we explore two additional ablations that determine the effects of state merging and the grammar rule exclusions presented in Table~\ref{tab:ode_systems}.

\begin{table}[]
\centering
\caption{Overview of methods included in our comparative study. Traits include whether finite differencing of state trajectories are required \textit{(F.D.)}, whether the method estimates all ODEs in a system jointly or independently \textit{(Joint Disc.)}, whether the method accounts for the flux-based nature of the dynamical systems \textit{(Fluxes)}, and what general category the method falls under.}
\label{tab:methods}
\begin{tabular}{@{}llllll@{}}
\toprule
\textbf{Method} & \textbf{F.D.} & \textbf{Joint Disc.} & \textbf{Fluxes} & \textbf{Category} & \textbf{Ref} \\ \midrule
Ours & \xmark & \cmark & \cmark & MCGS &  \\
Ours (No Stoic.) & \xmark & \cmark & \xmark & MCGS &  \\
SPL & \cmark & \xmark & \xmark & MCTS & \cite{sun2022symbolic} \\ 
SINDy & \cmark & \cmark & \xmark & Sparse regression & \cite{brunton2016discovering, pysindy} \\
PySR & \cmark & \xmark & \xmark & Genetic programming & \cite{cranmer2023interpretable} \\
ProGED & \xmark & \cmark & \xmark & Monte Carlo, grammar-based & \cite{omejc2024probabilistic} \\
ODEFormer & \xmark & \cmark & \xmark & Pretrained transformer & \cite{d2023odeformer} \\
\bottomrule
\end{tabular}
\end{table}

We evaluate each method using its default or recommended hyperparameters. 
To prevent a biased evaluation, we applied our framework's grammar rules to the grammar-based methods, SPL and ProGED. While PySR is not grammar-based, it allows the specification of operations and functions which we configured to match our grammar. We were, however, unable to apply our framework's associated grammar rule sampling probabilities to these alternative methods as they do not allow sampling at the flux-based level. 
We selected third-order polynomial features for SINDy's library of candidate functions, including square-root and rational terms for SIR and Fairen-Velarde respectively to better mimic these systems. Unfortunately, SINDy cannot multiply or nest features, so it is unable to fully reconstruct the square root SIR ODEs and the Fairen-Velarde ODEs using the available feature set. As a pre-trained, zero-shot inference method, ODEFormer required no additional specification.

\section{Results} \label{sect:results}

\subsection{Performance of FluxDisco}
\begin{figure}
    \centering
    \includegraphics[width=\linewidth]{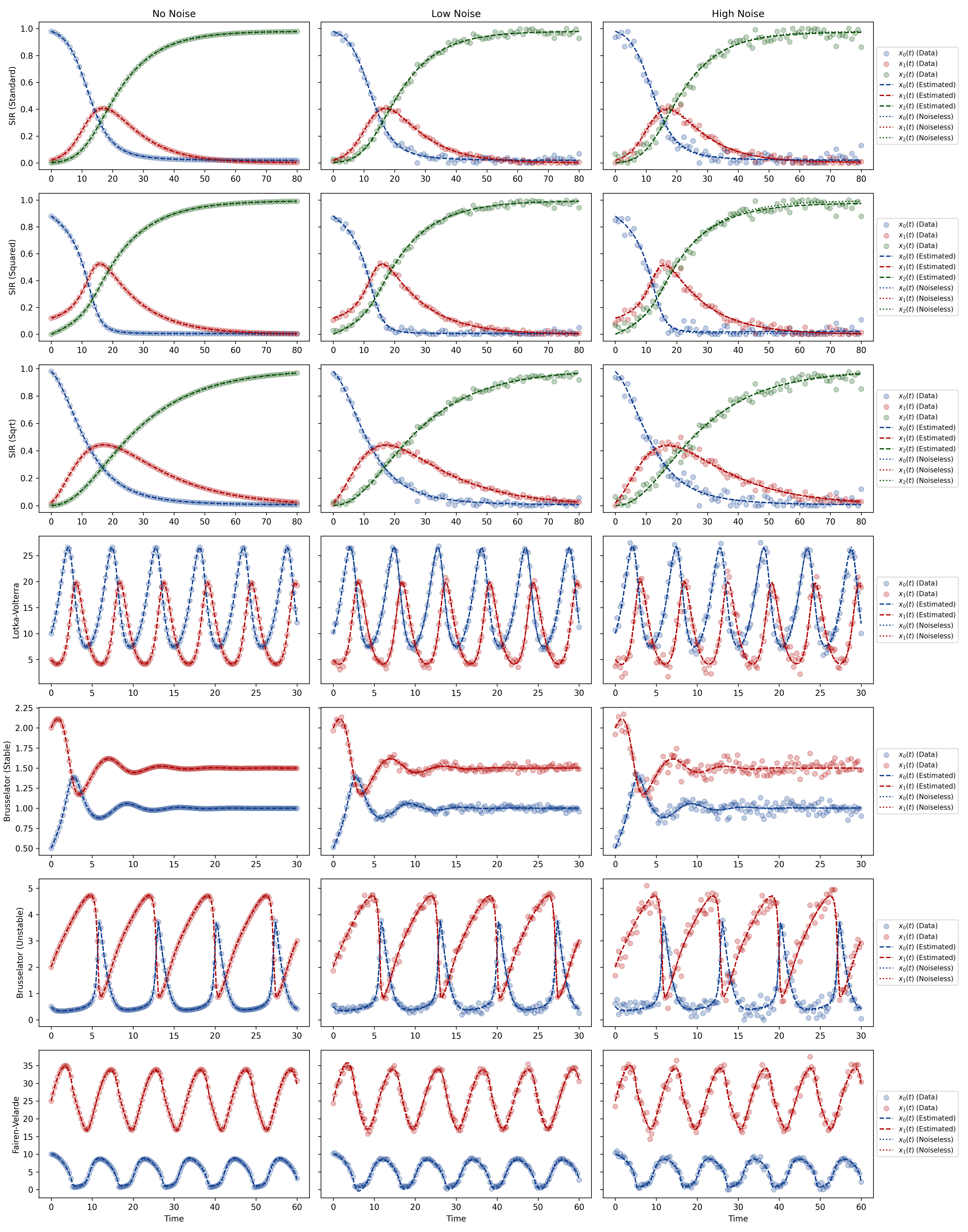}
    \caption{Comparison of estimated, noiseless, and observed trajectories.}
    \label{fig:all_trajectories}
\end{figure}

Figure~\ref{fig:all_trajectories} allows us to visually compare FluxDisco's estimated state trajectories against the noiseless and observed trajectories across all ODE systems considered in this study. Across all systems and regimes, the estimated trajectories closely align with the ground truth trajectories, suggesting that our method accurately recovers system dynamics, even in the presence of noise.

Our method also demonstrates high numerical accuracy across all systems, with the maximum observed trajectory reconstruction error remaining below 2\%.
\begin{figure}[htbp]
    \centering
    \includegraphics[width=\linewidth]{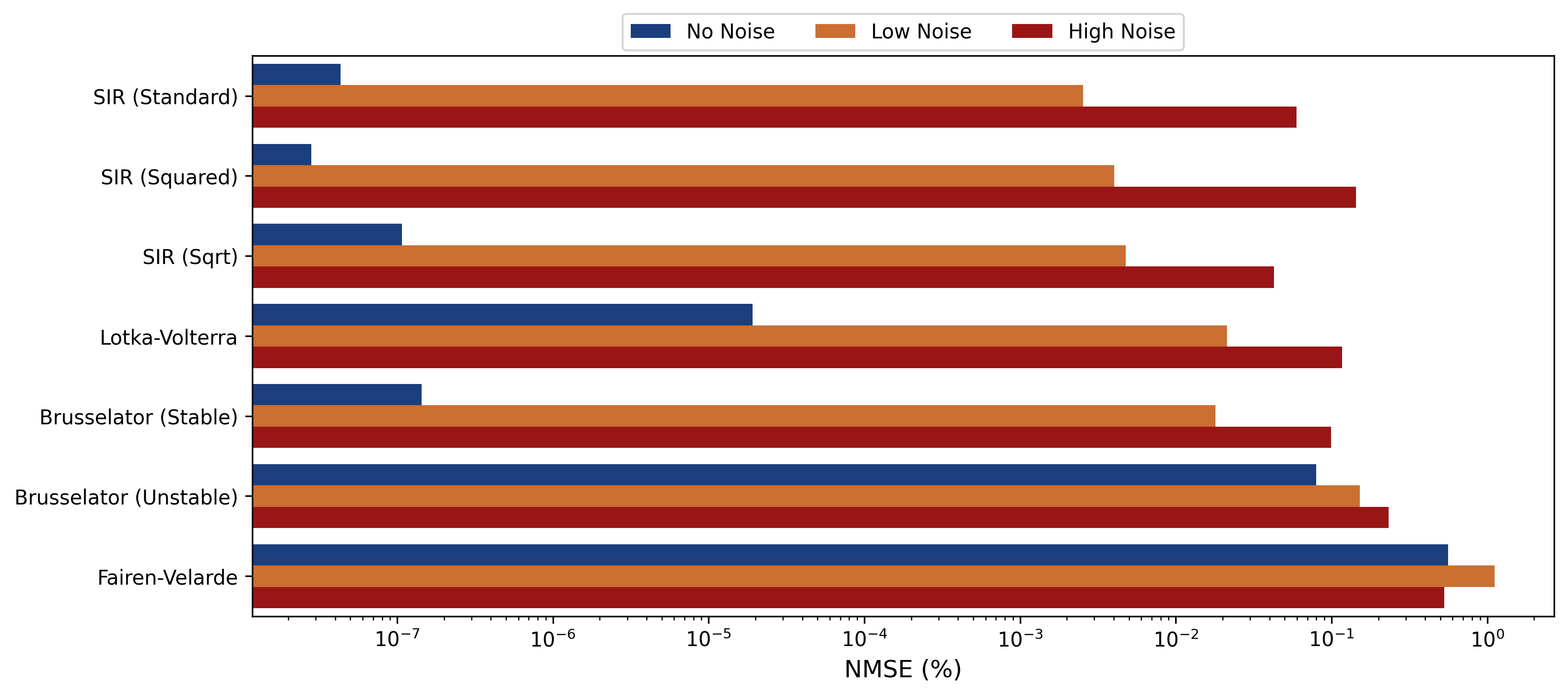}
    \caption{Normalised MSE (NMSE) across all benchmarked systems and levels of noise. Errors are averaged across fluxes per system and noise level.}
    \label{fig:nmse}
\end{figure}
As expected, Figure~\ref{fig:nmse} shows that as noise is introduced, the error generally increases. The Fairen-Velarde system produces the highest errors across all systems considered, reflecting the complex nature of its ODEs. After Fairen-Velarde, the unstable Brusselator yields the highest errors, likely due to the presence of both fast and slow dynamics.

The ablation studies in Section~\ref{sect:supp-ablation} of the Supplementary Materials illustrate that two of FluxDisco's key features, state merging and grammar rule exclusions, improve numerical accuracy and efficiency, but these gains diminish as the computational budget of the search increases. The results of the ablation without an imposed stoichiometry, shown in Section~\ref{sect:method-comp-results}, demonstrate that imposing known stoichiometries improves both numerical and symbolic performance, as well as computational efficiency.  

\subsection{Method Comparison} \label{sect:method-comp-results}
\begin{figure}
    \centering
    \includegraphics[width=\linewidth]{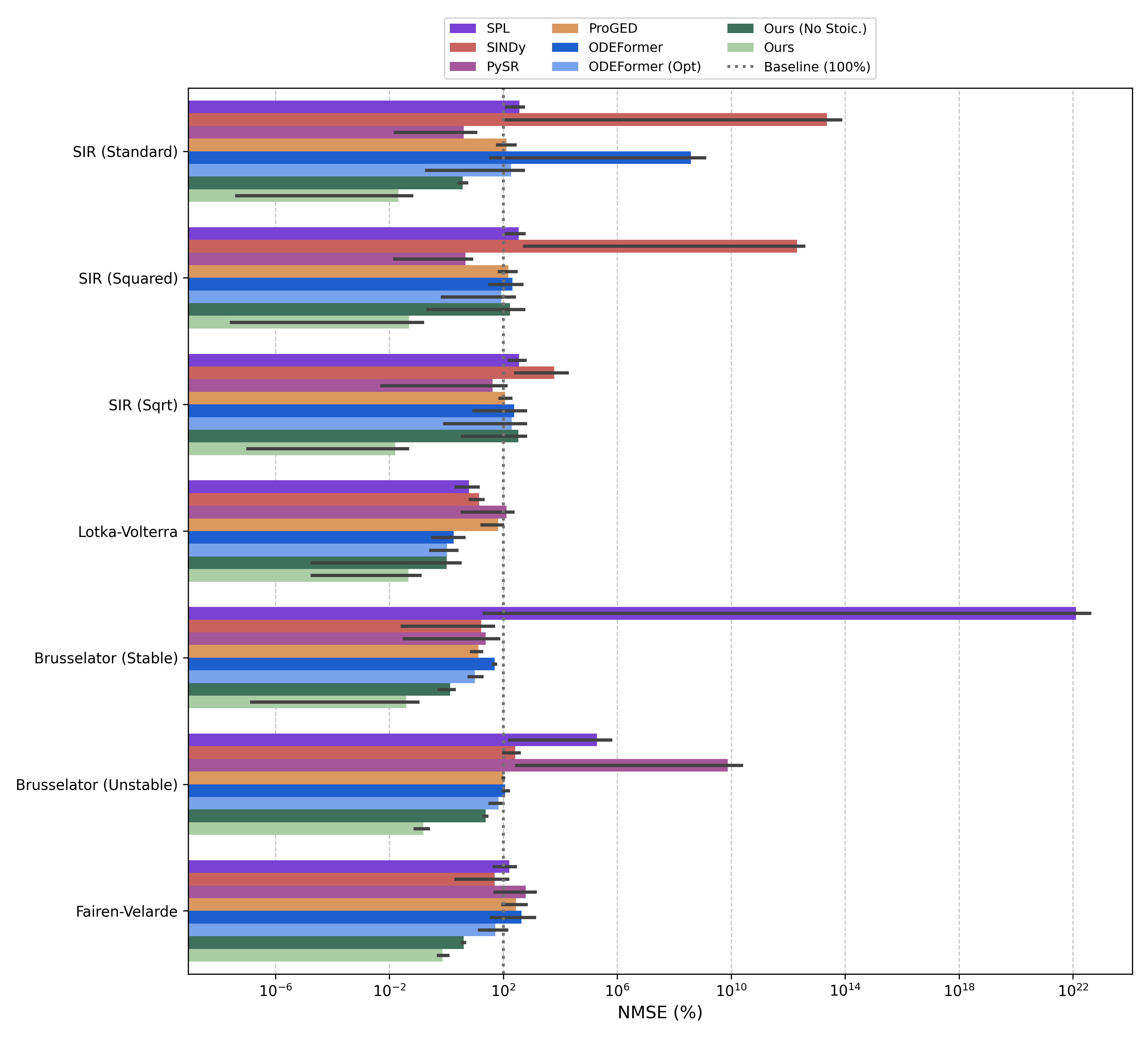}
    \caption{Comparison of Normalised Mean Squared Error (NMSE) across benchmarked models. 
    Error lines indicate the range of NMSE across varying noise levels. The vertical line denotes the baseline error associated with predicting the data mean across all time points (NMSE = 100\%).}
    \label{fig:nmse_models}
\end{figure}

\paragraph{Numerical Results}
Figure~\ref{fig:nmse_models} shows the NMSE for the benchmarked methods across all considered ODE systems and noise levels. The bars indicate the mean error across all noise levels, and the error bars represent the full range of these errors. 

Our method produces the lowest errors across all systems ($<2\%$), remaining well below the baseline error. The version of our framework which does not impose stoichiometry demonstrates considerably worse performance than the stoichiometry-aware variant, illustrating the benefit of enforcing known system structure in ODE discovery.
While we expect well-performing methods to yield errors lower than the na\"ive baseline, all methods besides ours produced errors greater than the baseline for some experiments. Some of the errors exceed the baseline by orders of magnitude, illustrating that these approaches are prone to divergent behaviour and fail to reconstruct system dynamics through their estimated expressions. 

\paragraph{Symbolic Results}
 
\begin{figure}[htbp]
    \centering
    \includegraphics[width=0.9\linewidth]{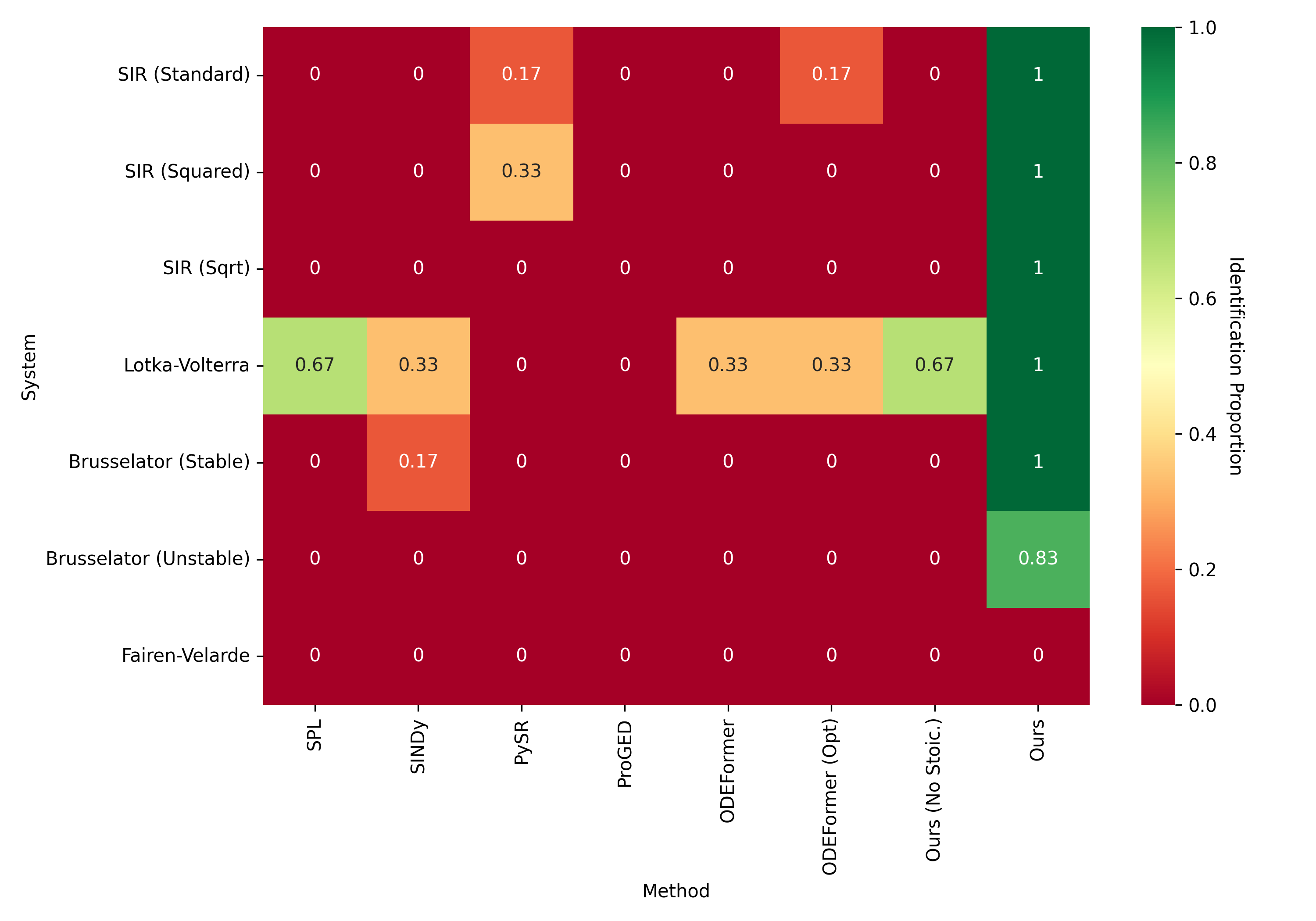}
    \caption{Coupled term identification accuracy in top-predicted equations across benchmarked methods for each ODE system. Results are averaged across all noise levels.}
    \label{fig:coupling_heatmap}
\end{figure}

Figure~\ref{fig:coupling_heatmap} presents the identification accuracy of the coupled expressions for all benchmarked methods. Our method achieves high accuracy across all systems except the Fairen-Velarde system which remains entirely unidentified by all methods. Our ablation without an imposed stoichiometry performs as poorly as the remaining benchmarked systems, demonstrating that enforcing known stoichiometries allows for better identification of coupled dynamics.

\begin{figure}[htbp]
    \centering
    \begin{subfigure}{\textwidth}
        \centering
        \includegraphics[width=0.9\linewidth]{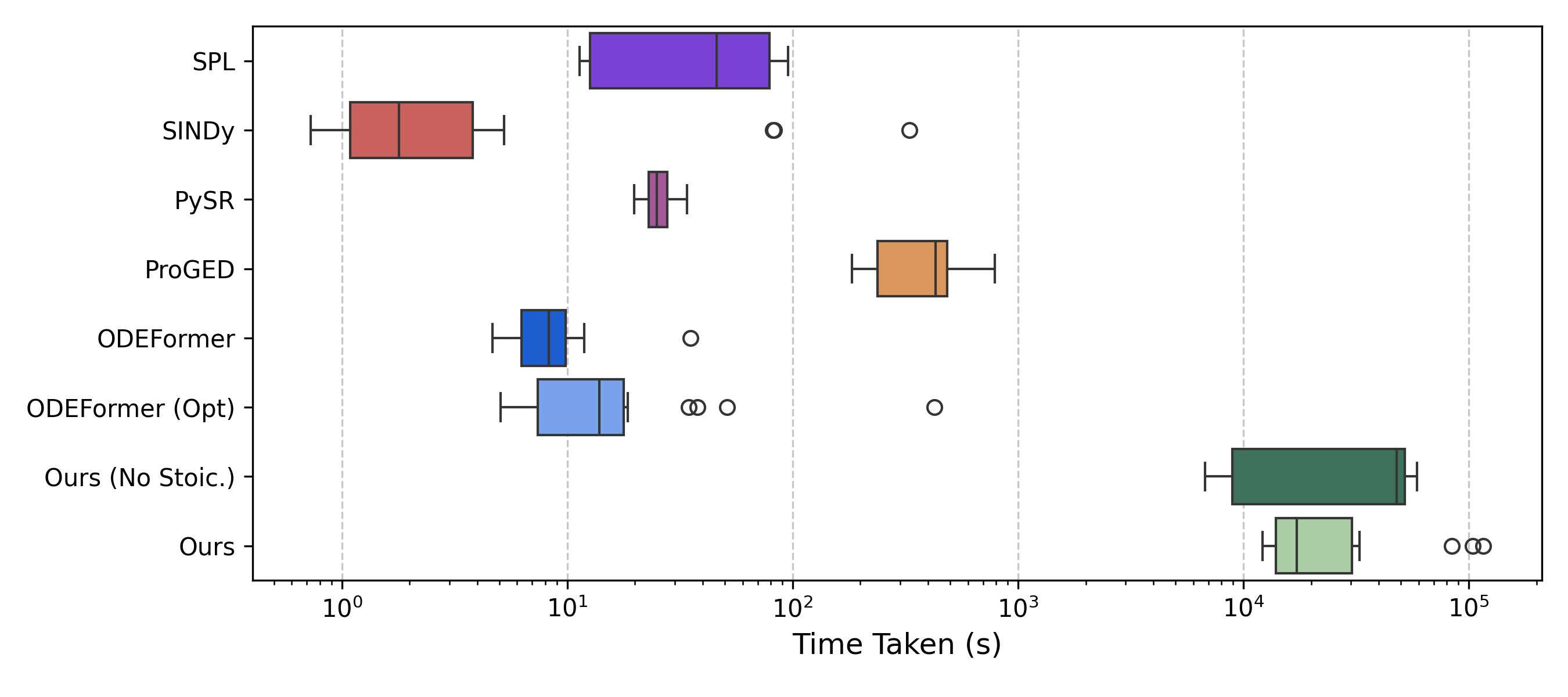}
        \caption{Runtime (seconds)}
        \label{fig:timing}
    \end{subfigure}
    \vspace{0.05em}
    
    \begin{subfigure}{\textwidth}
        \centering
        \includegraphics[width=0.9\linewidth]{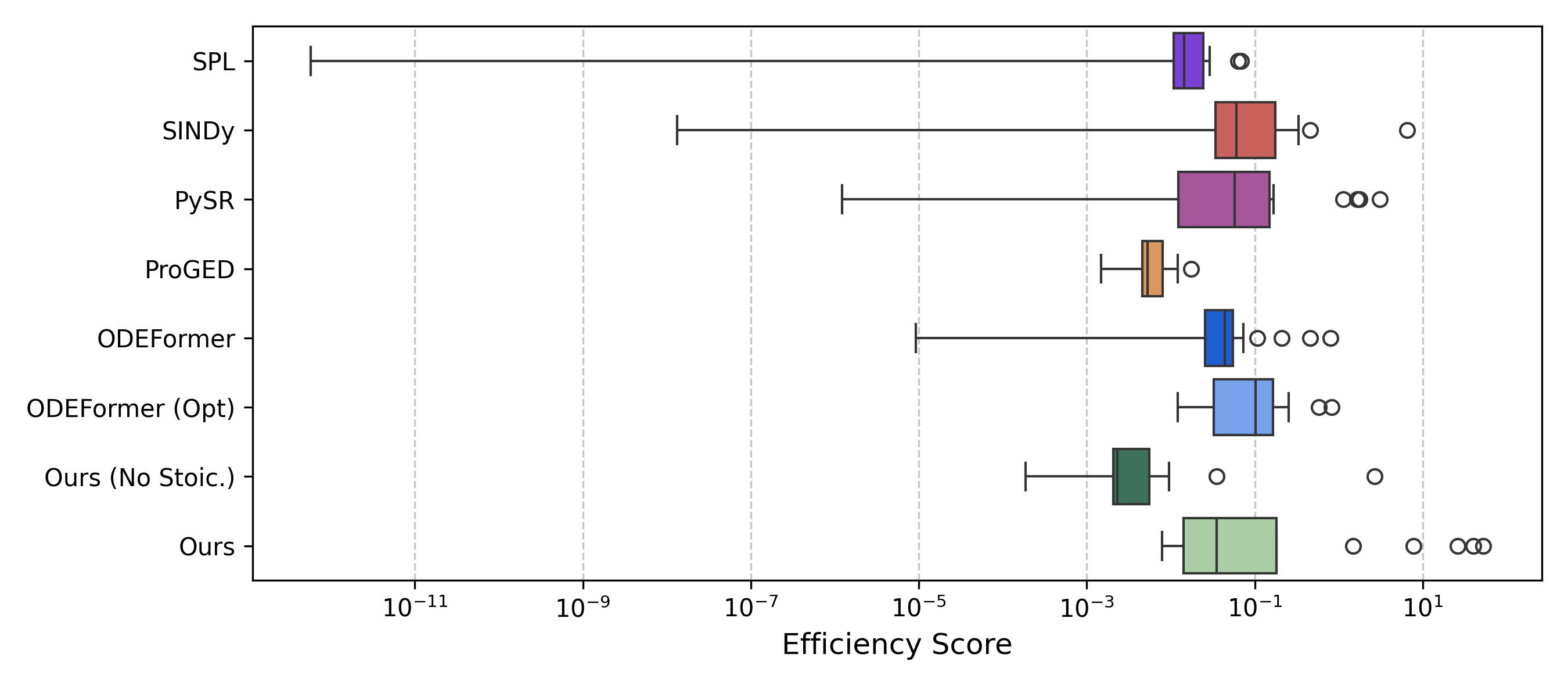}
        \caption{Efficiency score}
        \label{fig:efficiency}
    \end{subfigure}
    \caption{Comparison of runtime and efficiency across benchmarked models across ODE systems and noise levels. The efficiency score equally balances error (NMSE) and runtime (seconds).}
    \label{fig:time-comp}
\end{figure}

\paragraph{Efficiency Results}
Although our method achieves the lowest error and highest identification rate among the benchmarked methods (Figure~\ref{fig:nmse_models} and Figure~\ref{fig:coupling_heatmap}), this accuracy comes at the expense of computational speed. As shown in Figure~\ref{fig:timing}, the runtime for both variants of our method exceeds that of alternative methods by orders of magnitude.

Box-plots of the efficiency scores (Figure~\ref{fig:efficiency}) show that SPL, SINDy, PySR, and ODEFormer are highly volatile, with lower whiskers extending close to zero. While ProGED and the variant of our framework without imposed stoichiometry exhibit lower volatility, these methods are unable to achieve competitive efficiency scores. Our stoichiometrically-aware method and ODEFormer (Opt) exhibit both tight interquartile ranges and high efficiency scores. The discrepancy in efficiency between the two variants of our framework demonstrates that imposing a known stoichiometry successfully constrains the search space, allowing for better identification of dynamics.
Although ODEFormer (Opt) has a higher median efficiency score than our framework, our approach demonstrates a higher peak efficiency. We argue that the efficiency of ODEFormer and ODEFormer (Opt) shown in Figure~\ref{fig:efficiency} is overstated given that it does not account for the multi-day pre-training regime that the model underwent prior to inference. Should this pre-training be taken into account, the ODEFormer results would seem less competitive than our approach.

\section{Discussion}

In benchmarking our framework against a range of top-performing symbolic regression methods (Table \ref{tab:methods}), we find that our approach produces the lowest trajectory reconstruction errors and the highest identification rate of coupled dynamics. 
The benchmarked methods show inconsistent relative performance, although all fall short of our approach. While our framework yields the longest runtimes, its increased accuracy justifies this trade-off as equation discovery is typically an offline task and is rarely subject to strict runtime constraints.

To transparently establish the boundaries of our approach, we have included a challenging benchmark, the Fairen-Velarde system, which has comparatively complex dynamics. Our method discovered expressions for this system that produce well-fitting trajectories, but do not precisely match the ground truth expressions. This example highlights a common limitation of symbolic regression methods, with none of the benchmarked approaches successfully recovering its governing equations. Identifying exact mathematical expressions becomes increasingly difficult in complex systems, both due to an exploding combinatorial search space and the fact that different functional forms can produce similar trajectories. 

One of the greatest limitations of this work is its poor ability to scale to high-dimensional, complex ODE systems.
This drawback can largely be attributed to joint flux estimation. While critical for producing physically-adherent equations, joint estimation requires the algorithm produce the correct expressions across all fluxes within the same graph node. As the number of fluxes, or their complexity, increases, the probability of sampling a full set of correct expressions diminishes rapidly. This limitation is shared by most symbolic regression methods---as expression complexity or system dimensionality increases, the expression search space explodes. 

A further limitation of our work is that our method imposes a hard constraint on the system's stoichiometry. Consequently, stoichiometry misspecification will lead to inaccurate estimation of flux expressions. While our framework is designed for scenarios where the stoichiometry is known \textit{a priori}, this constraint can be relaxed. 
Firstly, as demonstrated in our ablation study, our method can be used without modification by setting the stoichiometry matrix to the identity matrix.
In this setting, the full system of ODEs become the target of estimation, though the coupling between the ODEs is lost and they are estimated independently.
Alternatively, our method could be modified to use a soft constraint, allowing both the stoichiometry and flux expressions to be estimated while penalising deviations from an expected stoichiometry within the reward function. 
To our knowledge, no existing methods for flux-based systems employ this type of soft constraint.
Existing literature focuses on estimating stoichiometry and flux expressions within the confines of mass action kinetics and low-order reactions \cite{hoffmann2019reactive, jiang2022identification, banos2026stoichiometrically, chen2025kinformer}.

\section{Conclusion}
This study addresses the challenge of extracting physically meaningful governing equations from noisy, multi-dimensional data in stoichiometric dynamical systems. Most existing dynamical symbolic regression methods are generalised and are therefore not suitable for the constraints associated with flux-based systems \cite{d2023odeformer, brunton2016discovering, sun2022symbolic, cranmer2023interpretable, omejc2024probabilistic}. 

To overcome these challenges and identify physically-adherent governing equations, we propose a method that jointly estimates the functional form of system fluxes while imposing a known system stoichiometry.
To do this, we adapt and specialise the MCGS algorithm \cite{leurent2020monte} for the purpose of joint flux discovery. 

We demonstrate the robustness of our method across a diverse set of physical systems and varying levels of observation noise. Our results illustrate that our proposed method, FluxDisco, accurately recovers governing dynamics where existing approaches fall short.

A promising avenue for future work is relaxing the assumption of full system observability. This assumption is often violated when considering data from real-world physical systems. Therefore, extending our framework to handle latent states would greatly improve its practical utility. 
This may involve latent state trajectory reconstruction using neural networks \cite{grigorian2025learning, liu2025structured}, or traditional approaches such as time-delay embeddings \cite{brunton2017chaos}. 
Furthermore, addressing the scalability challenges associated with joint flux estimation is critical for applying our method to high-dimensional, complex ODE systems.

\printbibliography[heading=bibintoc,title=References]

\newcommand{\beginsupplement}{
        \setcounter{table}{0}
        \renewcommand{\thetable}{S\arabic{table}}
        \renewcommand{\theHtable}{S\arabic{table}}
        
        \setcounter{figure}{0}
        \renewcommand{\thefigure}{S\arabic{figure}}
        \renewcommand{\theHfigure}{S\arabic{figure}}
        
        \setcounter{equation}{0}
        \renewcommand{\theequation}{S\arabic{equation}}
        \renewcommand{\theHequation}{S\arabic{equation}}
        
        \setcounter{section}{0}
        \renewcommand{\thesection}{S\arabic{section}}
        \renewcommand{\theHsection}{S\arabic{section}}
        
        \setcounter{algorithm}{0}
        \renewcommand{\thealgorithm}{S\arabic{algorithm}}
        \renewcommand{\theHalgorithm}{S\arabic{algorithm}}
}

\renewcommand{\floatpagefraction}{0.69}

\beginsupplement

\phantomsection 
\addcontentsline{toc}{section}{Supplementary Material}

\begin{center}
    {\huge Supplementary Material}
\end{center}
\vspace{1em}

\section{Methodology Details}
\label{sect:supp-search}
This section expands on core components of our methodology. Specifically, we justify our use of the stochastic reward variant of the Monte Carlo Graph Search (MCGS) algorithm (Section~\ref{sect:supp-det-vs-stoch}), detail our formulation of the mean reward credible intervals required in the stochastic reward setting (Section~\ref{sect:supp-reward-cis}), and present the complete reward calculation procedure (Section~\ref{sect:supp-reward-cal}).

\subsection{Deterministic vs Stochastic Rewards} \label{sect:supp-det-vs-stoch}
The MCGS algorithm supports both deterministic and stochastic reward settings \cite{leurent2020monte}. While either formulation can be applied to the task of symbolic regression, we opt to use the stochastic formulation.

In the deterministic setting of MCGS, rewards are only generated at terminal states corresponding to complete mathematical expressions. When flux expressions are partially-complete, we cannot evaluate the error between the partial expressions and the data. Therefore, non-terminal states receive a zero reward, resulting in a sparse-reward MDP. In such environments, an agent cannot easily distinguish between promising and unpromising partial expressions, resulting in an exhaustive exploration of the search space that scales poorly with ODE dimension and grammar size.

This inefficiency is compounded by the optimism inherent to the original MCGS algorithm \cite{leurent2020monte}.
The deterministic selection rule selects the next state $s'\in \mathcal{S}(s)$ according to:
\begin{align*}
    s \leftarrow \arg\max_{s' \in \mathcal{S}(s)} r(s') + \gamma \mathcal{U} (s')
\end{align*}
where $r(s')$ is the deterministic reward yielded by state $s'$ and $\mathcal{U}(s')$ is the upper bound on the value of state $s'$ based on Bellman optimality \cite{leurent2020monte}.
Because non-terminal states yield zero rewards, selection is driven solely by the optimistically initialised upper bounds of values of child states, $\mathcal{U}(s') \ \forall s'\in \mathcal{S}(s) $. The optimistic initialisation of unexplored nodes suppresses signals from discounted terminal states and prevents them from meaningfully guiding the search. Therefore, the search algorithm prioritises unexplored nodes, exploring the state space almost uniformly until a significant proportion of terminal states are reached. Reaching a large proportion of terminal states may be infeasible in large state spaces, such as the ones we are attempting to search. Without sufficient coverage of the state space, the search strategy remains undirected and uniform in deterministic settings.

To overcome the undirected nature of the search in sparse-reward settings, we employ the stochastic reward variant of MCGS (Main Text Section~\ref{sect:select}). Rather than assigning a zero reward to non-terminal nodes, the flux expressions are completed using a rollout policy to obtain an estimate of a node's potential reward. 
By introducing dense, sampled rewards, the agent can better discriminate between neighbouring nodes leading to a more efficient navigation of the search space. 

\subsection{Reward Confidence/Credible Intervals} \label{sect:supp-reward-cis}

The stochastic formulation of MCGS utilises confidence intervals for the mean empirical reward of all states $(l(s), u(s)) \forall s$ to account for reward stochasticity. 
Leurent and Maillard~\cite{leurent2020monte} construct these intervals using the Binary Kullback-Leibler (BKL) divergence.
The tightness of the BKL bounds depends on the ratio between the total number of rewards sampled across the graph and the number of local samples drawn. 
In large state-action spaces, states are rarely revisited so this ratio remains high. Therefore, the confidence bounds remain as initialised throughout the search, resulting in little differentiation between nodes during selection. 
Consequently, this formulation is overly conservative for equation discovery due to its expansive search space.

As introduced in the main text, we leverage Thompson sampling \cite{agrawal2012analysis} to overcome the conservatism of BKL bounds.
While Thompson sampling is designed for binary reward systems, the framework can be generalised to the scalar, bounded reward setting.
The scheme assumes an uninformative prior on the mean of the empirical reward distribution of a node.
When rewards are sampled, the prior is updated based on the Beta-Bernoulli conjugate prior:
\begin{align*}
    \text{Prior}: \Theta &\sim \text{Beta}(\alpha = 1, \beta = 1) \quad \textit{(uniform distribution)}\\
    \text{Likelihood}: (R | \Theta = \theta) &\sim \text{Bernoulli}(\theta) \\
    \text{Posterior}: (\Theta|R = \mathbf{r}) &\sim \text{Beta}\left(\alpha=1+\sum_{i=1}^n r_i,\ \beta = 1 + n - \sum_{i=1}^n r_i \right)
\end{align*}
When rewards are scalar, we use a heuristic to maintain the Bayesian update framework. A scalar reward $r_i \in [0,1]$ can be considered as $r_i$ units of success and $1-r_i$ units of failure. Therefore, when a reward $r_i$ is sampled, the success parameter $\alpha$ of the posterior distribution can be incremented by the reward, $r_i$, and the failure parameter can be incremented by $1-r_i$. The resulting posterior distribution is:
\begin{align}
    \text{Beta}\left(\alpha = 1 + \sum_{i=1}^n r_i, \beta = 1 + \sum_{i=1}^n (1-r_i) \right) \equiv
    \text{Beta}(\alpha = 1 + n\bar{r}, \beta = 1 + n(1-\bar{r}))
\end{align}
While the conjugate prior framework technically assumes integer parameters, due to the Bernoulli likelihood function, this relaxation serves as a useful heuristic to derive a pseudo posterior distribution for the mean reward.
The credible intervals $(l(s), u(s))$ are then extracted via the quantiles of this distribution, as detailed in Equation~\ref{eq:beta_credible_intervals} of the main text.

The primary motivation for this approach is that it does not rely on the ratio between the total number of node visits across the graph to the current node's visit count, improving its suitability for large search spaces. Furthermore, the Beta credible intervals are immediately responsive to sampled rewards, whereas the BKL intervals require a significant number of samples before they respond to the reward signal. 

\subsection{Reward Calculation} \label{sect:supp-reward-cal}

Algorithm \ref{alg:reward_calc} details the complete procedure for evaluating the reward associated with a sampled set of candidate flux expressions post-rollout.

\begin{algorithm}
\caption{Reward calculation} \label{alg:reward_calc}
\begin{algorithmic}[1]
    \Require 
    \Statex \textbf{Candidate flux expressions:} $\hat{\mathbf{v}}(\mathbf{x}(t)) = [\hat{v}_0(\mathbf{x}(t)), \dots, \hat{v}_{F-1}(\mathbf{x}(t))]$
    \Statex \textbf{Initial conditions:} $\mathbf{x}(0)=[x_0(0), \dots, x_{D-1}(0)]$
    \Statex \textbf{Stoichiometry matrix:} $\mathbf{S} \in \mathbb{R}^{D \times F}$ 
    \Statex \textcolor{lightgray}{\rule{\linewidth}{0.2pt}}
    \vspace{0.01em}
    \Statex \hspace*{-\algorithmicindent} \textbf{Algorithm:}
    \State \textbf{Constant folding:} Apply constant folding to candidate fluxes $\hat{\mathbf{v}}(\mathbf{x}(t))$ to obtain simplified expressions, $\hat{\mathbf{v}}_{\text{fold}}(\mathbf{x}(t))$. 
    \State \textbf{Compute complexity:} Calculate the system complexity $\mathcal{C}$ from the folded expressions.
    \State \textbf{Apply stoichiometry:} Join the folded flux expressions using the stoichiometry matrix: 
    \[
    \hat{\dot{\mathbf{x}}}(t) = 
    \mathbf{S} \hat{\mathbf{v}}_{\text{fold}}(\mathbf{x}(t))
    \]
    \Repeat
        \State \textbf{Propose and apply constants:} Propose constants $\mathbf{c}$ via optimisation algorithm and substitute them into the system equations, $\hat{\dot{\mathbf{x}}}(t)$.
        \State \textbf{Solve ODEs:} 
        \[ \hat{\mathbf{x}}(t) = \text{ODESolver}\left(\hat{\dot{\mathbf{x}}}(t),\ \mathbf{x}(0)\right) \]
        \State \textbf{Compute optimisation error:} Derive the sum of squared errors (SSE) between estimates and observed data to guide constant optimisation:
        \[
            \text{SSE}(\hat{\mathbf{x}}(t), \mathbf{x}(t)) = \sum_{d =0}^{D-1} \sum_{n=1}^N \sum_{i=1}^T \left(\hat{x}_{d}^{(n)}(t_i) - x_{d}^{(n)}(t_i)  \right)^2
        \]

    \Until{Convergence met or max iterations reached.}

    \State \textbf{Compute total error:} Using the optimised constants, compute total error.
    \[ \text{MSE}_{\text{total}} = \frac{\text{SSE}(\hat{\mathbf{x}}(t), \mathbf{x}(t))}{N \times T} \]

    \State \textbf{Compute reward:} 
    \[ R = \eta^\mathcal{C} \exp \left(- \frac{\text{MSE}_{\text{total}}}{\tau} \right)\]
    \vspace{0.7em}
     \hspace*{-\algorithmicindent} \Return Reward $R$, folded flux expressions $\hat{\mathbf{v}}_{\text{fold}}(\mathbf{x}(t))$, optimised constants $\mathbf{c}$.
\end{algorithmic}
\end{algorithm}

\section{Hyperparameters}
The hyperparameter choices for our framework, FluxDisco, are outlined in Table~\ref{tab:hyperparams}. 

\begin{table}[h!]
\centering
\caption{Selected hyperparameter values}
\label{tab:hyperparams}
\begin{tabular}{@{} l p{5.5cm} l p{5.5cm} @{}}
\toprule
\textbf{Symbol} & \textbf{Description} & \textbf{Value} & \textbf{Fairen-Velarde Modifications} \\ \midrule
$\gamma$ & Discount factor for future rewards & 0.9 &  \\ \addlinespace
$\epsilon$ & Accuracy measure for propagation (Algorithm 4, \cite{leurent2020monte}) & 0.01 &  \\ \addlinespace
$\tau$ & Temperature in reward function & 0.005 & Increased to 0.05 to increase reward scale. \\ \addlinespace
$\eta$ & Parsimony coefficient in reward function & 0.99 & Increased to 0.995 to allow for more complex expressions. \\ \addlinespace
$\kappa$ & Maximum flux expression depth & 6 & Increased to 12 to allow for more complex expressions. \\ \addlinespace
 & Episodes & 100 & Decreased to 40 to compensate for increased expression depth. \\ \addlinespace
 & Max iterations per episode & $4\kappa+2$ &  \\ \addlinespace
 & Rollouts per node post-selection & 1 &  \\ \addlinespace
 & Warm start rollouts per node & 2 &  \\ \addlinespace
 & Number of data simulations, or realisations, per ODE system & 1 &  \\ \bottomrule
\end{tabular}
\end{table}

\section{Data Simulation} \label{sec:supp-data}
The data used in this study are simulated. Algorithm~\ref{alg:data_sim} details the simulation process, and Table~\ref{tab:data-params} outlines the simulation parameters of each benchmarked ODE system.
\begin{algorithm}
\caption{Data Simulation} \label{alg:data_sim}
\begin{algorithmic}[1]
    \Require 
    \Statex \textbf{Noise level:} $\epsilon$
    \Statex \textbf{Ground truth ODEs:} $\dot{\mathbf{x}}(t)$
    \Statex \textbf{Initial conditions:} $\mathbf{x}(0)=[x_0(0), \dots, x_{D-1}(0)]$
    \Statex \textbf{Constant values:} $\mathbf{c}$
    \Statex \textbf{ODE solver:} ODESolver
    \Statex \textcolor{lightgray}{\rule{\linewidth}{0.2pt}}
    \vspace{0.05em}
    \Statex \hspace*{-\algorithmicindent} \textbf{Algorithm:}
    \State \textbf{Get noiseless trajectories:} Solve specified ODEs:
    \[
    \mathbf{x}(t) = \text{ODESolver}(\dot{\mathbf{x}}(t), \mathbf{x}(0))
    \]
    \If{$\epsilon > 0$}
    \State \textbf{Calculate variable ranges:} For each system variable $d \in \{0, \dots, D-1\}$, compute the range:
    \[
    \Delta x_d = \max_t(x_d(t)) - \min_t(x_d(t)) \quad \forall d
    \]
    \State \textbf{Generate Gaussian noise:} Randomly sample noise from the standard normal distribution to produce $\mathbf{Z} \in \mathbb{R}^{T \times D}$ where each element is
    \[
    z_{[t, d]} \sim \mathcal{N}(0, 1) \quad \textit{(i.i.d.)}
    \]
    \State \textbf{Scale and apply noise:} Apply scaled noise to the noiseless trajectories:
    \begin{align*}
        \tilde{x}_d(t) &= x_d(t) + \epsilon \cdot \Delta x_d \cdot z_{[t, d]} \quad \forall t, d
        \\
        \tilde{\mathbf{x}}(t) &= [\tilde{x}_0(t), \dots, \tilde{x}_{D-1}(t)]
    \end{align*}
    \State \textbf{Ensure non-negativity:} The systems considered \textit{(SIR, Lotka-Volterra, Brusselator, and Fairen-Velarde)} cannot take on negative values, so correct for any negativity induced by noise:
        \[
        \tilde{\mathbf{x}}(t) = \max(\tilde{\mathbf{x}}(t), \textbf{0})
        \]
    \If{system is SIR}
        \State \textbf{Normalise proportions:} Normalise data within each time step to ensure variables sum to 1:
        \[
        \tilde{x}_d(t)=\frac{\tilde{x}_d(t)}{\sum_{d=0}^{D-1}\tilde{x}_d(t)} \quad \forall t, d
        \]
    \EndIf
    \ElsIf{$\epsilon=0$}
        \State \textbf{Return clean trajectories:} If no noise is specified, return the noiseless trajectories:
        \[
        \tilde{\mathbf{x}}(t) = \mathbf{x}(t)
        \]
    \EndIf
    
     \hspace*{-\algorithmicindent} \Return trajectories $\tilde{\mathbf{x}}(t)$.
\end{algorithmic}
\end{algorithm}

\begin{table}[h!]
\caption{Simulation parameters for the benchmarked ODE systems, including initial conditions $\mathbf{x}(0)$, constant values $\mathbf{c}$, time spans, sampling frequencies, and noise levels $\epsilon$.}
\label{tab:data-params}
\begin{tabular}{@{}llllll@{}}
\toprule
\textbf{System} & $\mathbf{x}(0)$ & $\mathbf{c}$ & \textbf{Time Span} & \textbf{Time Points} & \textbf{Noise} \\ \midrule
SIR (Standard) & $[0.98, 0.02, 0]$ & $[0.4, 0.1]$ & $[0, 80]$ & $80$ & $[0, 0.02, 0.05]$ \\
SIR (Squared) & $[0.88, 0.12, 0]$ & $[1.3, 0.08]$ & $[0, 80]$ & $80$ & $[0, 0.02, 0.05]$ \\
SIR (Square root) & $[0.98, 0.02, 0]$ & $[0.15, 0.06]$ & $[0, 80]$ & $80$ & $[0, 0.02, 0.05]$ \\
Lotka-Volterra & $[10, 5]$ & $[1, 0.1, 1.5]$ & $[0, 30]$ & $150$ & $[0, 0.03, 0.07]$ \\
Brusselator (Stable) & $[0.5, 2]$ & $[1.5, 1]$ & $[0, 30]$ & $150$ & $[0, 0.03, 0.07]$ \\
Brusselator (Unstable) & $[0.5, 2]$ & $[0.5, 2]$ & $[0, 30]$ & $150$ & $[0, 0.03, 0.07]$ \\
Fairen-Velarde & $[10, 25]$ & $[0.5, 15, 10]$ & $[0, 60]$ & $150$ & $[0, 0.03, 0.07]$ \\ \bottomrule
\end{tabular}
\end{table} 

Because the ODE systems considered have unit constraints, we need to process the noisy trajectories to ensure these constraints are adhered to. All benchmarked systems require observations to be non-negative, therefore any negative values caused by the applied noise are set to zero. While this processing step introduces a slight positive bias, it is necessary to produce data that is physically realistic. The SIR epidemic model has an additional unit constraint. The system variables represent proportions and should sum to one. Therefore, we normalise the data per time step after zeroing any negative values to meet this extra constraint. This normalisation step introduces additional bias by correlating the noise across the system variables.

Table~\ref{tab:data-params} shows that the number of simulated time points and noise levels for the SIR system deviate from the remaining systems. For the SIR systems, initial conditions and constants values were selected to ensure that across all regimes, infections rose to similar peaks around approximately the same time period for fair comparison. Based on these parameters, a time span of $[0, 80]$ days was chosen to ensure that a full epidemic is realised while avoiding an excessively long tail. While the other dynamical systems are simulated across 150 time points, the epidemic models are restricted to 80 time points. This choice reflects the practicalities of epidemiological data collection, where data is typically reported at most daily. 
The noise levels for the SIR system were reduced to compensate for the lower sampling frequency and the bounded data range. SIR trajectories lie close to the boundaries of the data range for extended periods, therefore they are very susceptible to the biases introduced by the post-processing, unit constraint corrections. Lowering the noise for this system mitigates the effect of this bias.

Whilst our framework allows for multiple realisations from the dynamical systems, we focus on the more challenging single-realisation setup, using only a single initial condition per system.

\section{Ablation Studies} \label{sect:supp-ablation}
To evaluate the efficacy of key features of FluxDisco and validate our design decisions, we conduct various ablation studies. 
We demonstrate the impact of imposing known stoichiometries in Section~\ref{sect:results} of the main text, where the ablation without imposed stoichiometries shows markedly worse numerical and symbolic performance.
Here, we present two additional ablation studies that isolate the effects of state merging, and excluding grammar rules by zeroing select sampling probabilities. 

\subsection{State Merging}

To evaluate whether state merging improves symbolic regression performance, we compare our method, FluxDisco, against an ablation which does not merge states, structuring the search space as a tree as opposed to a graph. To assess performance relative to the search budget, we recorded key metrics every five episodes. 

\begin{figure}[p]
    \centering
    \includegraphics[width=\linewidth]{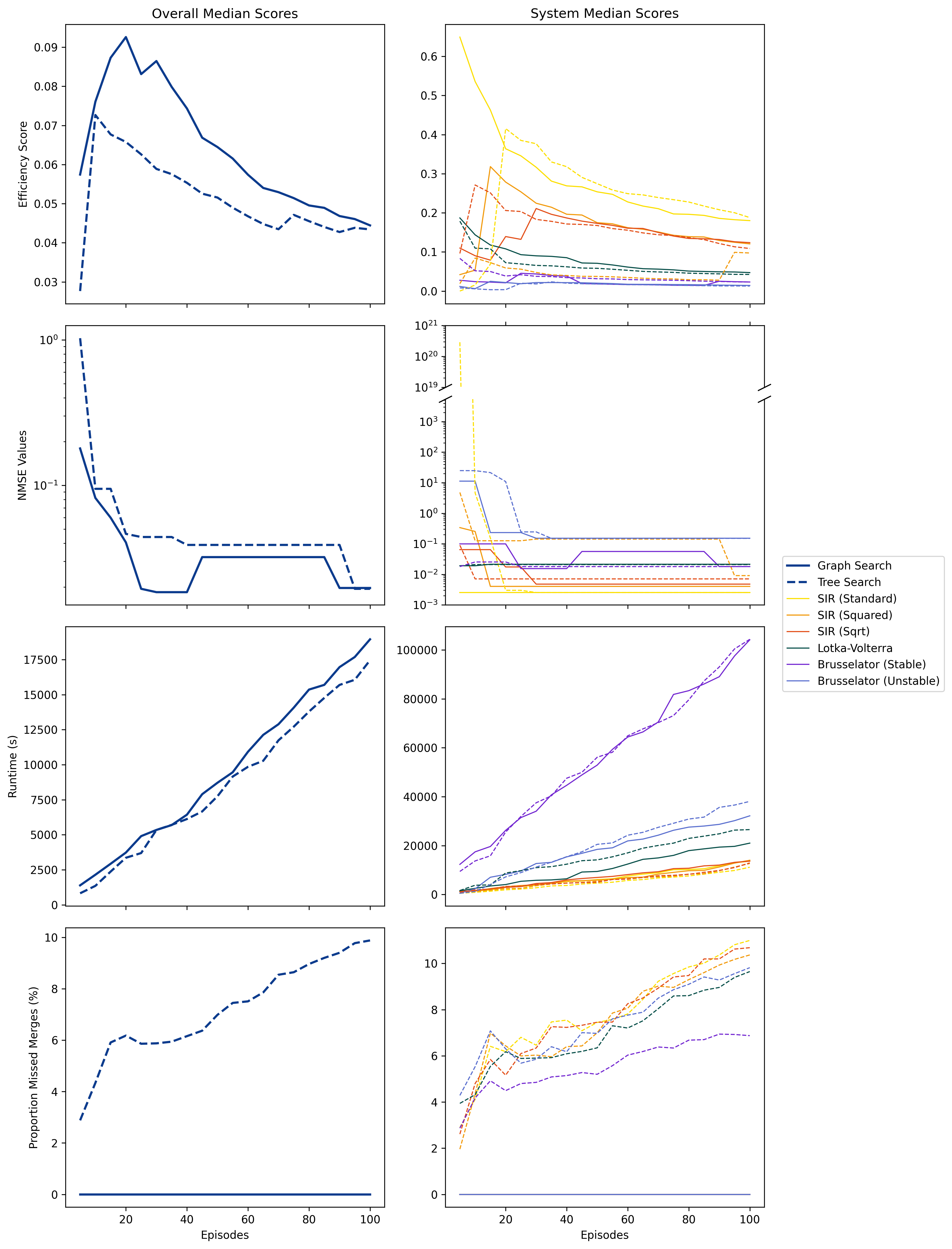}
    \caption{Graph search vs.~tree search ablation study results. The \textit{left} panels show median scores aggregated across all conditions, and the \textit{right} panels break down median scores by ODE system.}
    \label{fig:merging_metrics}
\end{figure}

Figure~\ref{fig:merging_metrics} presents our results, evaluating performance across four key metrics: efficiency score, normalised mean squared error (NMSE), computational runtime, and the proportion of missed state merges.
For this analysis, we exclude results from the Fairen-Velarde system. 
The Fairen-Velarde system is our most challenging benchmark with complex dynamics necessitating deeper search structures. To compensate for the increased computational demands of this search, fewer episodes were conducted for this system (Table~\ref{tab:hyperparams}). Since the Fairen-Velarde system has unique hyperparameters, including episode count, it is excluded from this ablation study to ensure a uniform comparison.

Our graph search method achieves higher median efficiency scores and lower median errors (NMSE) than the tree search ablation without state merging. 
While there is stochastic variation in the individual ODE system results (Figure~\ref{fig:merging_metrics}, right column), the aggregated median scores (Figure~\ref{fig:merging_metrics}, left column) illustrate a clear trend. 
The performance gap is most noticeable at lower computational budgets (episode counts), where the improved sample efficiency and wider information propagation of our graph-based method allow for faster discovery of high-quality solutions.
This advantage is supported by the steep increase in the proportion of missed state merges by the tree search method in early stages of the search. 
As the computational budget increases, the median performance of the two methods converge, illustrating that both methods find good solutions when provided sufficient search budget. 

While state merging introduces overheads that increase runtime, the difference in median runtimes is marginal. Runtimes for the individual ODE systems vary substantially and in some cases, the long-run tree search runtimes exceed those of the graph searches (e.g. Lotka-Volterra, unstable Brusselator). 

\subsection{Grammar Rule Exclusions}
\begin{figure}
    \centering
    \includegraphics[width=\linewidth]{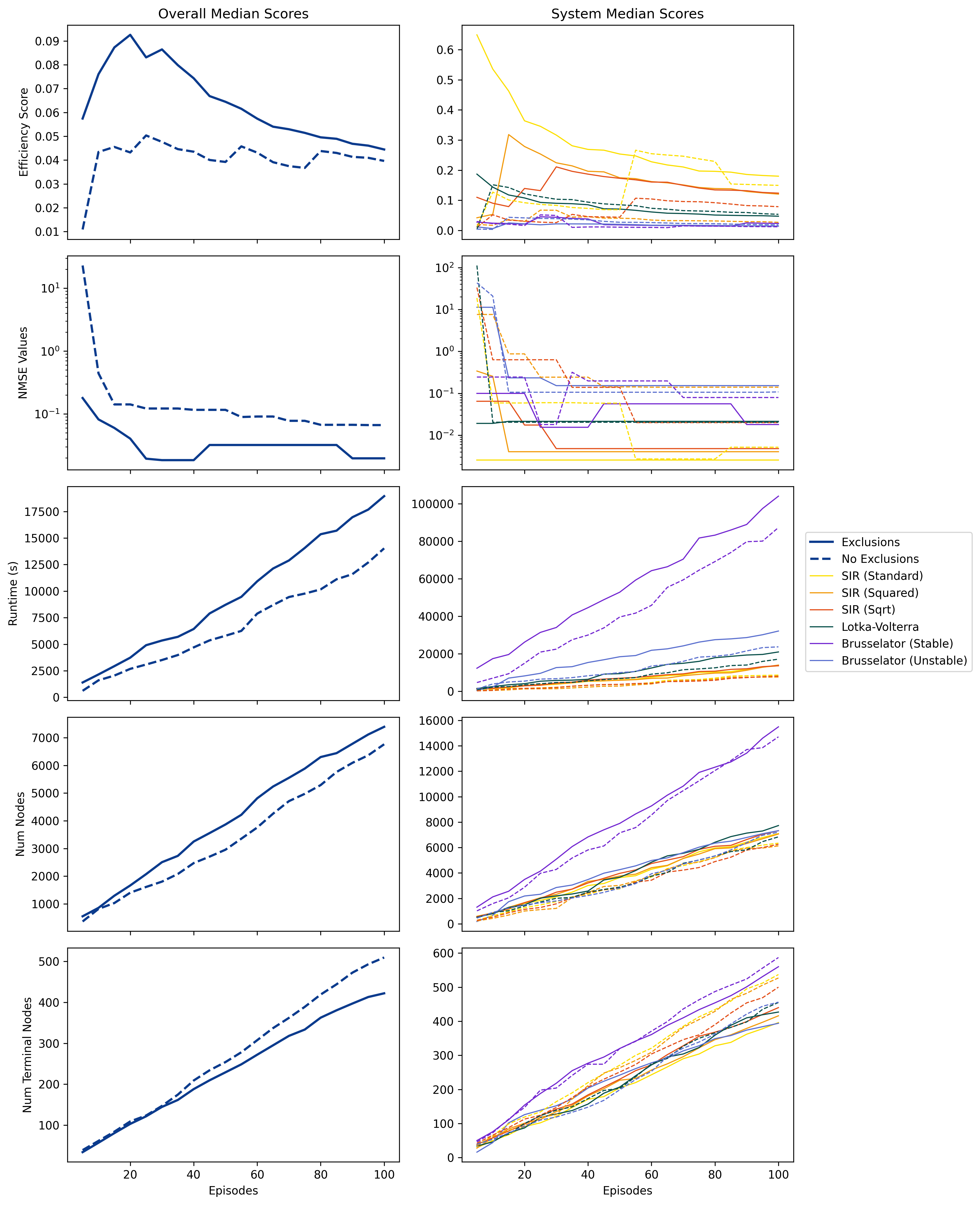}
    \caption{Search results with grammar rule exclusions vs.~ablation without exclusions. The \textit{left} panels show median scores aggregated across all conditions, and the \textit{right} panels break down median scores by ODE system.}
    \label{fig:exlusion_ablation}
\end{figure}

Our framework is physics-informed in that we enforce known system structures by constraining system stoichiometry. Furthermore, we allow prior domain knowledge to be incorporated through user-specified grammar rule sampling probabilities (Section~\ref{sect:rollout}). In the main text, we zero specific sampling probabilities for fluxes from the SIR, Lotka-Volterra, and Brusselator systems to reflect prior physical knowledge. To determine the effect of these exclusions, we run an ablation study using uniform sampling probabilities across all defined grammar rules.

Figure~\ref{fig:exlusion_ablation} compares the search performance with and without grammar rule exclusions. The excluded grammar rules, shown in Table~\ref{tab:ode_systems} of the main text, all correspond to terminal actions. Consequently, the ablation results produce graphs with more terminal nodes and shorter selection paths. Since search episodes terminate once either a terminal set of equations are constructed, or a maximum number of iterations are reached, having more terminal actions will terminate episodes earlier, reducing the overall runtime. Conversely, excluding these terminal grammar rules should yield longer selection paths, larger graphs with more nodes, and comparatively fewer terminal nodes.

While the ablation corresponds to faster runtimes due to a higher density of terminal nodes, its efficiency scores are lower than the main method's scores due to poorer candidate solutions being identified. However, as the search progresses, the ablation is able to find solutions with sufficiently low errors causing the efficiencies of both approaches to converge.

\end{document}